\documentclass[]{imag-ms-template}

\title{A Multimodal Sequence-to-Sequence Model for Cross-Subject Prediction of Brain Responses to Naturalistic Stimuli} 

\author{
Qianyi He,$^{1\ast}$ Monica D. Rosenberg,$^{2,3,4}$ Yuan Chang Leong$^{2,3,4\ast}$\\
{\small $^{1}$Department of Neuroscience, City University of Hong Kong, Hong Kong SAR, China}\\
{\small $^{2}$Department of Psychology, University of Chicago, Chicago, IL, USA}\\
{\small $^{3}$Neuroscience Institute, University of Chicago, Chicago, IL, USA}\\
{\small $^{4}$Institute for Mind and Biology, University of Chicago, Chicago, IL, USA}\\
{\small $^\ast$Correspondence: qianyihe@cityu.edu.hk; ycleong@uchicago.edu}
}

\date{}

\begin{document} 

\maketitle 

\begin{abstract}

Brain encoding models predict time-resolved neural activity from computational representations of ongoing experience, providing a principled framework for testing how information is represented and transformed across cortical systems. Naturalistic audiovisual narratives are a particularly rich but challenging testbed for these models, requiring integration of multimodal inputs over long temporal horizons and generalization across individuals with substantial response variability. We introduce a multimodal sequence-to-sequence Transformer with a hybrid cross-subject parameterization that predicts cortex-wide parcel-wise fMRI time series autoregressively from visual, audio, language, and vision--language representations. We evaluate the approach on data from the Courtois NeuroMod project, where four deeply-sampled participants viewed six seasons of \textit{Friends} and four feature-length films during fMRI. Sequence-to-sequence temporal modeling yields consistent improvements over single-frame prediction across cortical networks, with gains extending to novel stimuli. A hybrid architecture that pairs a shared stimulus encoder with lightweight subject-specific decoder components outperforms both fully shared and fully individual models, indicating complementary advantages of learning shared stimulus representations across subjects and fitting individual neural readouts. Finally, we show that in data-scarce settings, hybrid models can be personalized to new individuals with limited fMRI data, demonstrating that multi-subject pretraining serves as a strong inductive prior for building individual-specific encoding models.  Together, these results indicate that combining multimodal sequence modeling with a hybrid cross-subject architecture offers a scalable framework for personalized brain encoding under naturalistic conditions.

\end{abstract}

\keywords{neural encoding models, naturalistic fMRI, multimodal integration, sequence modeling, cross-subject generalization}

\section{Introduction}

Naturalistic movie paradigms offer an ecologically valid window into human cognition, where visual information from the video stream, acoustic information from the soundtrack, and linguistic information derived from dialogue unfold in parallel and interact over extended temporal horizons \citep{sonkusare2019naturalistic,finn2022naturalistic}. A growing body of work in cognitive and systems neuroscience aims to build brain encoding models that predict time-resolved neural activity from such continuous, multimodal experience \citep{tang2023brain,oota2025multi,dong2023modeling}. Accurate encoding models provide a direct, quantitative mapping  from stimulus representations to neural responses, making it possible to test which computational features explain activity across cortical regions \citep{naselaris2011encoding,huth2016natural}. In the context of naturalistic stimuli, such models can shed light on the neural processes underlying cross-modal integration, hierarchical processing, and the distribution of information across large-scale networks \citep{khosla2021cortical,takagi2025cross,li2024enhancing}. Beyond their explanatory value, accurate encoding models also serve as a foundation for improved brain decoding, as increased model-brain alignment strengthens downstream stimulus reconstruction and decoding pipelines that rely on reliable mappings between stimuli and brain responses \citep{vu2011encoding}.

Recent years have seen a rapid growth of dense neuroimaging datasets collected while participants view naturalistic movies and other audiovisual narratives, often with repeated or extended viewing sessions spanning diverse content within the same individuals \citep{boyle2023courtois,hanke2016studyforrest,pinho2024individual}. These resources make it increasingly feasible to train more expressive (i.e., non-linear, context-dependent) whole-brain encoding models and to systematically evaluate their generalization across stimulus domains or tasks. The encoding models applied to these data have generally followed a standard pipeline: extracting features from continuous stimuli, concatenating them over a short temporal window, and mapping them to voxel-wise brain responses using regularized linear regression, often implemented as finite impulse response (FIR) models estimated via ridge regression \citep{dupre2025voxelwise}. Within this framework, the models learn linear mappings that predict each timepoint in the fMRI signal independently from recent stimulus history. 

While this approach has proven successful in many settings, it typically does not explicitly model temporal dependencies in measured brain responses, treats different stimulus modalities independently rather than modeling their interactions, and is generally fit to each subject independently. As a result, the models make poor use of structure shared across subjects and require a large amount of individual data. Here, we propose a sequence-to-sequence Transformer-based architecture to better capture the temporal dependencies and multimodal interactions that shape fMRI signals over time. Similar to neural network-based machine translation models that map sequences of words
from one language to another \citep{sutskever2014sequence}, our model translates sequences of multimodal stimulus representations into sequences of whole-brain neural responses, conditioning each prediction on both the full input sequence and the history of prior neural activity.

Beyond stimulus–response modeling, whole-brain generalization across individuals remains a central barrier for naturalistic encoding models. Even after anatomical normalization, fine-grained functional topographies vary substantially across people, so naively pooling subjects can blur shared structure and undercut prediction \citep{frost2012measuring}. Prior work has addressed this through functional alignment methods such as hyperalignment \citep{haxby2020hyperalignment} and the shared response model (SRM) \citep{chen2015reduced}, which project neural responses onto a common space via subject-specific transformations.  However, these alignment steps are typically fit separately from the encoding model and therefore not optimized for prediction. An alternative strategy is to fit or fine-tune models to each individual subject \citep{freteault2025alignment}, which can improve personal fit, but increases the amount of subject-specific data required and can be difficult to estimate reliably in low-data regimes. 

Recent work on cross-subject transfer and constructing personalized encoding models from small datasets suggests a useful compromise: learn a strong shared representational scaffold from pooled data while adapting only a lightweight subject-specific mapping \citep{wen2018transferring, gu2022personalized,feilong2023individualized,tang2025semantic}. Guided by the intuition that stimulus representations may be broadly shared across people, whereas each person’s brain response to those representations can differ, we adopt a hybrid architecture with a shared encoder and subject-specific components within the decoder. This design combines the benefits of pooled training with the ability to capture individual differences. We further integrate hybrid parameterization, multi-step prediction and multimodal integration within a single sequence-to-sequence encoding framework.

We develop and evaluate this framework using the Algonauts 2025 dataset from the Courtois NeuroMod project, in which four deeply-sampled participants viewed six seasons of {Friends} and four feature-length films during fMRI \citep{gifford2024algonauts}. This dataset provides a natural testbed for evaluating generalization, as models trained on {Friends} can be tested on held-out episodes (i.e., in-distribution generalization), or on the feature-length films, which differ substantially in characters, genre, pacing, and visual style (i.e., out-of-distribution generalization). We previously developed an initial version of our cross-subject multimodal model as an entry to the Algonauts 2025 challenge, which aimed to predict fMRI responses to these naturalistic stimuli. As required by the organizers, the entry was released as a non-peer-reviewed preprint  \citep{he2025multimodal}. 

In the present study, we systematically evaluate the benefits of sequence modeling and hybrid cross-subject parameterization to encoding model performance. First, we ask whether sequence-to-sequence temporal modeling yields reliable gains over single-timepoint prediction under both in-distribution and out-of-distribution evaluation, and whether these gains vary systematically across cortical networks. Second, we test whether a hybrid architecture that combines a shared encoder with lightweight subject-specific  decoder components consistently outperforms models that are either fully shared across subjects or fit to each subject independently. Third, we assess how efficiently the model can be personalized to new individuals from limited fMRI data by measuring few-shot adaptation performance. To further ground these findings in practical scenarios, we compare different parameter-sharing strategies across transfer settings that vary in both data availability and distributional shift. Together, these findings advance our understanding of how temporal dependencies and the balance between shared and individual components of neural representation can be leveraged to build encoding models that generalize across both stimulus domains and individuals.

\section{Methods}
\subsection{Dataset}
\subsubsection{Participants and stimuli}
All analyses were conducted on the publicly released Algonauts 2025 Challenge dataset, curated from the Courtois NeuroMod (CNeuroMod) project. The release includes the original audiovisual movie stimuli, TR-aligned stimulus annotations for multimodal feature extraction, time-stamped dialogue transcripts, and standardized fMRI derivative time series for model development and evaluation \citep{gifford2024algonauts, boyle2023courtois}. We used the four participants with extensive naturalistic-viewing fMRI data, identified in the released dataset as sub-01, sub-02, sub-03, and sub-05.

The stimulus set comprises long-form audiovisual narratives spanning approximately 65 h, including six seasons of the TV show \textit{Friends} and four feature-length films from the CNeuroMod Movie10 dataset (\textit{The Bourne Supremacy (abbreviated as Bourne)}, \textit{Hidden Figures (Figures)}, the BBC documentary \textit{Life (Life)}, and \textit{The Wolf of Wall Street (Wolf)}). To keep individual fMRI acquisition runs at a tolerable duration, both episodes and films were divided into contiguous clips, and adjacent clips included a brief temporal overlap to preserve narrative continuity. During acquisition, visual stimuli were projected onto a screen in the MRI room and audio was delivered using MRI-compatible in-ear headphones with additional passive noise attenuation.

\subsubsection{MRI acquisition and preprocessing}
All fMRI data were acquired within the CNeuroMod project at the Functional Neuroimaging Unit (UNF) in Montréal on a Siemens Prisma Fit 3T scanner with a 64-channel head/neck receive coil. Head motion was minimized using individualized, custom-milled head cases. Functional images were collected with an accelerated simultaneous multi-slice gradient-echo EPI sequence (multiband factor 4; TR = 1.49\,s, TE = 37\,ms, flip angle = 52$^\circ$, 2\,mm isotropic voxels). Each session included a short reversed phase-encoding acquisition to enable retrospective susceptibility-distortion correction. High-resolution anatomical scans were acquired periodically to support alignment and spatial normalization. More details can be found in the \href{https://docs.cneuromod.ca/en/latest/MRI.html}{CNeuroMod MRI documentation} (https://docs.cneuromod.ca/en/latest/MRI.html).

For all analyses, we used the parcel-wise BOLD time series distributed with the Algonauts 2025 dataset. These time series are normalized to MNI space and summarized into 1,000 functionally defined parcels based on the Schaefer atlas \citep{schaefer2018local}. The underlying CNeuroMod source data follow the BIDS convention \citep{gorgolewski2016brain} and were preprocessed using fMRIPrep \citep{esteban2019fmriprep}. We used the released parcel-wise data as provided and did not perform additional image-space preprocessing.

\subsubsection{Data splits and evaluation settings}
\label{Methods splits}

For the primary model comparisons, data from \textit{Friends} Seasons 1--5 were used for model training. Episodes 1--12 of Season 6 were used for validation and hyperparameter selection, and Episodes 13--24 of Season 6 were held out for in-distribution (ID) testing. This split evaluates generalization to unseen episodes drawn from the same television series and therefore sharing similar characters, narrative structure, and audiovisual statistics with the training data.

All runs from the four feature-length films---\textit{Bourne}, \textit{Figures}, \textit{Life}, and \textit{Wolf}---were excluded from training and validation in the primary analyses and reserved for out-of-distribution (OOD) evaluation. These films differ from \textit{Friends} in genre, characters, pacing, cinematography, and narrative structure, providing a more stringent test of generalization to novel stimulus domains. The assignment of episodes and runs to each split was fixed across participants to ensure that model comparisons were based on the same stimulus content.

The primary split was used for the comparisons of temporal prediction architectures and cross-subject parameter-sharing schemes. The limited-data transfer analyses used separate, scenario-specific combinations of film and \textit{Friends} data, as described in Section~\ref{Methods:Cross subject transfer under data scarcity}. No test stimulus from a given transfer scenario was included in that scenario's training or validation data.

For each cortical parcel, we quantified accuracy using the Pearson correlation between predicted and observed fMRI on valid frames. Valid frames were defined by excluding the first 5 TRs of each run, which convey limited stimulus-locked information due to the latency of the BOLD response, and the final 5 TRs to minimize end effects arising from slight acquisition overrun, timing imprecision near stimulus offset, and the absence of corresponding stimulus features. Evaluation was restricted to time points for which model predictions were available. 

For \textit{Friends}, held-out time points were concatenated across episodes within each split before computing one Pearson correlation per parcel. For \textit{Movies}, split segments were concatenated within each film, and parcel-wise correlations were computed separately for each film. All correlations were transformed using Fisher's z transformation, with film-level z values first averaged across films for each subject in the movie analyses. Overall performance was computed by averaging z values across parcels and subjects. For network-level analyses, Fisher-transformed correlations were averaged across parcels within each network. All summary statistics are reported as Pearson r values after inverse Fisher transformation.

\subsection{Stimulus Feature Extraction}
Stimuli were provided as audiovisual clips (.mkv) together with time-stamped dialogue transcripts (.tsv). To align stimulus features with fMRI sampling (TR = 1.49 s), we discretized the continuous stimulus streams into non-overlapping 1.49-s bins and extracted one feature vector per bin for each feature stream. Feature streams were aligned by TR index, yielding synchronized time series for video, audio, language, and a cross-modal vision-language representation.

\textbf{Visual features.} 
Visual features were extracted with VideoMAE \citep{tong2022videomae}, a masked video autoencoder pretrained on large-scale video datasets, which provides spatiotemporal representations sensitive to motion dynamics and scene-level transitions. For each TR bin, we uniformly sampled 16 frames from the corresponding TR interval and encoded them jointly with VideoMAE. We used the resulting [CLS] representation from the final encoder layer as the TR-level visual embedding, yielding a 768-dimensional feature vector per TR. To incorporate longer-range visual context while preserving this TR-resolved sampling, we augmented each embedding with a cumulative-average context term by computing, at time $t$, the running mean of all previous TR-level embeddings within the same stimulus segment and concatenating it with the current embedding. This produced a 1{,}536-dimensional visual feature vector per TR and serves as a lightweight memory-like summary of preceding visual context via temporal smoothing, which is well suited to continuous narrative stimuli.

\textbf{Audio features.}
Audio features were extracted with HuBERT \citep{hsu2021hubert}, a self-supervised model that learns hierarchical speech representations
through masked prediction of latent units. For each TR bin, we extracted the corresponding audio segment from the stimulus clip, resampled it to 16 kHz, and converted it to a single-channel waveform by averaging channels. We then encoded each TR-aligned audio segment with HuBERT and obtained a TR-level representation by mean-pooling the hidden states over time. To capture complementary low-level and higher-level acoustic structure, we concatenated mean-pooled embeddings from the 3rd and 9th transformer layers \citep{vaidya2022self}, yielding a 1{,}536-dimensional audio feature vector per TR. 


\textbf{Language features.} Language features were derived from the dialogue transcripts using Qwen1.5-0.5B \citep{bai2023qwen}. For each TR bin, we constructed a textual context by concatenating the current TR text with all preceding transcript text within the same stimulus segment. The context was tokenized and truncated to the most recent 2{,}048 tokens to preserve recent narrative continuity within a fixed context length. From the hidden states of the 12th transformer layer \citep{bonnasse2024fmri}, we computed two complementary pooled representations: a history-level summary obtained by mean pooling token embeddings across the contextual sequence, and a current-focused summary obtained by mean pooling the embeddings of the last 10 tokens. Each pooling produced a 1{,}024-dimensional vector, and their concatenation yielded a 2{,}048-dimensional language feature vector per TR. If no transcript text was available up to a given TR bin, we assigned a zero vector.

\textbf{Joint visual-language features.} To complement the unimodal features, we used BridgeTower \citep{xu2023bridgetower}, a pretrained cross-modal transformer designed to integrate visual and textual signals. At each TR bin, the model received a representative video frame together with the corresponding TR-aligned utterance from the transcript, with empty text used when no dialogue was present. We extracted a 1{,}536-dimensional fused representation from the final-layer \texttt{pooler\_output}, which corresponds to the [CLS] embedding after vision--language fusion. These features provide additional cross-modal alignment that is difficult to obtain from independent visual and language encoders and complement the modality-specific representations.

\subsection{Model Architecture}
\label{Methods: Model Architecture}
\begin{figure}[htbp]
\begin{center}
\includegraphics[width=1.04\textwidth]{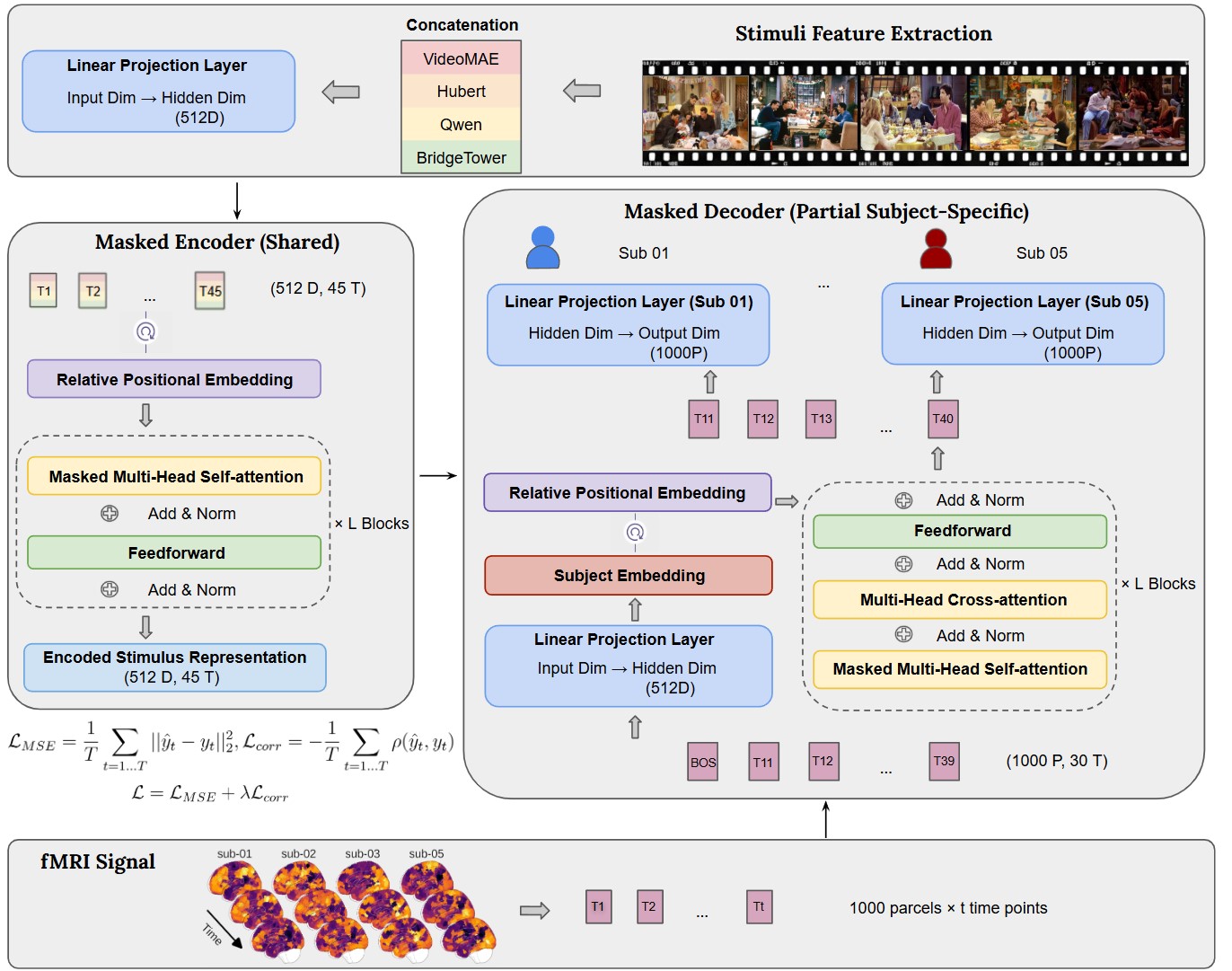}
\caption{\textbf{Overview of the proposed framework.} Multimodal stimulus features are extracted with pretrained models (VideoMAE, HuBERT, Qwen, and BridgeTower), concatenated, and linearly projected to a 512-dimensional sequence. A shared causal Transformer encoder maps the stimulus sequence to an encoded representation, and a masked autoregressive Transformer decoder predicts parcel-wise fMRI time series via cross-attention to this representation. To accommodate inter-subject variability, the decoder includes partially subject-specific components. The model is trained by minimizing $\mathcal{L}=\mathcal{L}_{\mathrm{MSE}}+\lambda\,\mathcal{L}_{\mathrm{corr}}$. See Methods \ref{Methods: Model Architecture} for more details.} 
\label{structure}
\end{center}
\end{figure}

We propose a Transformer-based encoder-decoder architecture that maps sequences of multimodal stimulus features to sequences of fMRI responses. The design accounts for both the sequential dynamics of neural activity and inter-subject variability. Departing from conventional static TR-wise regression approaches, our model predicts fMRI time series in an autoregressive manner, employing masked self-attention and cross-modal contextual integration over the multimodal stimulus features. The overall framework is shown in Fig.~\ref{structure}.

The encoder is implemented as a Transformer with causal self-attention, such that each time step only attends to its current and preceding inputs. This departs from the standard bidirectional design of sequence-to-sequence (seq2seq) Transformers used in machine translation and is motivated by the nature of neural processing: the brain receives external stimuli sequentially as they unfold over time, without access to future information. By imposing this temporal constraint, the encoder better reflects the sequential nature of biological perception. Before the encoder stack, we added a relative positional encoding to the input sequence, allowing the model to incorporate temporal order information. Let $\mathbf{H}_\text{enc} = \mathrm{Encoder}(\mathbf{X}) \in \mathbb{R}^{T \times d}$ denote the output of the encoder, where $T$ is the number of time steps and $d$ is the hidden dimension. 

The decoder is a masked causal Transformer that generates fMRI signals autoregressively. At each decoding step $t$, the decoder receives as query the embedding of the previously generated output $\hat{\mathbf{y}}_{t-1}$ and attends only to preceding decoder states through masked self-attention, preventing access to future fMRI outputs. The decoder additionally cross-attends to the encoded stimulus sequence \(H_{\mathrm{enc}}\), allowing it to learn flexible temporal correspondences between stimulus representations and BOLD responses.

Formally, the decoder's output at time $t$ is computed as:
\begin{equation}
\begin{aligned}
\tilde{\mathbf{y}}_t &= \hat{\mathbf{y}}_{t-1} \mathbf{W}_{\text{dec}} + \mathbf{b}_{\text{dec}} \\
\mathbf{z}_t^{(0)} &= \tilde{\mathbf{y}}_t + \mathrm{RelPosEnc}(t) \\
\mathbf{z}_t^{(l,1)} &= \mathrm{LayerNorm}\left( \mathbf{z}_t^{(l-1)} + \mathrm{SelfAttn}(\mathbf{z}_{\leq t}^{(l-1)}) \right) \\
\mathbf{z}_t^{(l,2)} &= \mathrm{LayerNorm}\left( \mathbf{z}_t^{(l,1)} + \mathrm{CrossAttn}_{\text{stim}}(\mathbf{z}_t^{(l,1)}, \mathbf{H}_{\text{enc}}) \right) \\
\mathbf{z}_t^{(l)} &= \mathrm{LayerNorm}\left( \mathbf{z}_t^{(l,2)} + \mathrm{MLP}(\mathbf{z}_t^{(l,2)}) \right) \\
\hat{\mathbf{y}}_t &= \mathbf{z}_t^{(L)} \mathbf{W}_{\text{out}} + \mathbf{b}_{\text{out}}
\end{aligned}
\end{equation}
All attention modules are multi-head and include dropout and residual connections. 

The training objective combines mean squared error (MSE) and negative Pearson correlation between the predicted and ground truth sequences:
\begin{equation}
    \mathcal{L}_{\text{MSE}} = \frac{1}{T} \sum_{t=1}^T \| \hat{\mathbf{y}}_t - \mathbf{y}_t \|_2^2, \quad 
\mathcal{L}_{\text{corr}} = -\frac{1}{T} \sum_{t=1}^T \rho(\hat{\mathbf{y}}_t, \mathbf{y}_t),
\quad 
\mathcal{L} = \mathcal{L}_{\text{MSE}} + \lambda \mathcal{L}_{\text{corr}}
\end{equation}
where $\rho$ denotes the Pearson correlation across cortical parcels at each time point, and $\lambda$ is a weighting factor.

To support modeling multiple subjects with the same model, we introduced a hybrid architecture in which the encoder is shared across all subjects, while the decoder incorporates individual-specific components. This design reflects the assumption that, even under identical stimuli, fMRI responses exhibit substantial inter-subject variability. At the same time, there is shared structure in how different individuals process the same stimuli, and leveraging data from multiple subjects allows the model to learn these common representations, improving model robustness and generalization. 

Each subject $s$ is associated with a learnable embedding vector $\mathbf{e}_s$ that is added to every time step of the decoder input. In addition, the final linear projection layer of the decoder head is specialized per subject, allowing personalized mapping from hidden representations to predicted fMRI outputs. This setup assumes that how stimuli are represented can be shared across individuals, while the corresponding brain responses remain subject-dependent. Formally, for subject $s$, two modifications were made to the original architecture:
\begin{equation}
\begin{aligned}
\mathbf{z}_t^{(0)} &= \tilde{\mathbf{y}}_t + \mathbf{e}_s + \mathrm{RelPosEnc}(t) \\
\hat{\mathbf{y}}_t^{(s)} &= \mathbf{z}_t^{(L)} \mathbf{W}_{\text{out}}^{(s)} + \mathbf{b}_{\text{out}}^{(s)} \quad (\text{for subject } s)
\end{aligned}
\end{equation}
Notably, this architecture enables parameter-efficient adaptation to new subjects by fine-tuning only the subject-specific embeddings and output head, making it suitable for few-shot personalization in fMRI encoding tasks.

\subsection{Sequence construction and training}

To increase the amount of supervision available for sequence-level fMRI prediction, we segmented each continuous stimulus--response sequence into overlapping temporal windows. Let $\mathbf{X}=[\mathbf{x}_1,\dots,\mathbf{x}_T]$ denote the multimodal stimulus feature sequence, and $\mathbf{Y}=[\mathbf{y}_1,\dots,\mathbf{y}_T]$ denote the corresponding parcel-wise fMRI response sequence. For each valid start time $t$, we constructed a training pair by extracting a stimulus segment of length $L_{\mathrm{stim}}$ and pairing it with an fMRI segment of length $L_{\mathrm{fmri}}$, with a temporal offset $\Delta$ between their starting points. Formally, the resulting training pair is defined as
\begin{equation}
\mathbf{X}^{(t)}=\mathbf{x}_{t: (t+L_{\mathrm{stim}}-1)},
\qquad
\mathbf{Y}^{(t)}=\mathbf{y}_{(t+\Delta): (t+\Delta+L_{\mathrm{fmri}}-1)}.
\end{equation}

Consecutive windows are sampled with stride $s$ to control overlap. This windowing strategy substantially increases the number of training samples and provides repeated supervision for individual fMRI time points under partially overlapping temporal contexts. During training, we shuffled the order of windowed samples at the beginning of each epoch to reduce temporal correlation among adjacent overlapping windows. This increases the diversity of temporal contexts in each mini-batch and improves the stability of stochastic optimization. During evaluation, overlapping predictions for the same TR were averaged. The temporal window configuration was selected using the validation procedure described in Section~\ref{Methods: Temporal model selection}.

For autoregressive decoding, we initialized the decoder with a learnable [BOS] token and employed the teacher forcing strategy to guide sequence generation. At each decoding step $t$, the model received either the ground-truth output $\mathbf{y}_{t-1}$ or its own previous prediction $\hat{\mathbf{y}}_{t-1}$ as input, selected according to a teacher forcing ratio $\gamma \in [0, 1]$. We linearly annealed $\gamma$ from 1.0 toward 0 over training epochs to gradually transition from fully guided training to self-conditioned generation. By reducing the influence of unreliable model predictions early in training, this schedule limits error accumulation and stabilizes autoregressive optimization. During validation and testing, generation was fully autoregressive, without access to ground-truth BOLD responses. Unless otherwise specified, all Transformer models used two encoder and two decoder layers with four attention heads, a hidden dimension of 512, and a dropout rate of 0.1. We optimize models using Adam with a learning rate of $1\times 10^{-4}$ and a batch size of 32 on NVIDIA A100 GPUs.

\subsection{Baseline Models}

We evaluate a set of baseline models to assess the contributions of explicit sequence modeling and cross-subject parameterization. For sequence modeling, we compare models that predict a single fMRI frame from a fixed stimulus context, including ridge regression and a Transformer sequence-to-one (seq2one) model, against models that predict a contiguous fMRI sequence, including our autoregressive seq2seq model. For cross-subject parameterization, we compare  training a single model pooled across all subjects, training a separate model for each subject independently, and our hybrid model, which shares the encoder across subjects while retaining subject-dependent components in the decoder. Unless otherwise specified, all baseline models use the same multimodal stimulus features, data splits, and evaluation metric.

\subsubsection{Baseline models for explicit sequence modeling}

We evaluated the effect of explicit sequence modeling by comparing three models that share the same multimodal stimulus features and data splits, but differ in whether they model fMRI responses as an autoregressive sequence or as independent targets. Ridge and Transformer seq2one predict a single fMRI frame from a stimulus context window, whereas the proposed Transformer seq2seq generates a contiguous fMRI subsequence. All models in this subsection were fit separately for each subject.

\textbf{Ridge.} We used ridge regression to predict a single fMRI frame from the recent stimulus history using a finite impulse response (FIR) design. This approach is widely used in fMRI encoding studies and serves as a linear baseline against which to evaluate the nonlinear Transformer-based models. Stimulus features at each TR were z-scored using training-set statistics and reduced to 512 dimensions using PCA fit on the training data only, with PCA parameters estimated in a streaming fashion over training episodes. For a context length $k$, the model input was constructed by concatenating the PCA features from the previous $k$ TRs to form a lagged design vector, which was paired with a single target BOLD frame at a fixed temporal offset. The regularization parameter was selected separately for each participant on the validation split.

\textbf{Seq2one.} To distinguish nonlinear temporal integration of the stimulus input from autoregressive sequence prediction, we trained a Transformer that maps a stimulus context window to a single fMRI frame. The model receives the most recent $k$ stimulus features and directly regresses one target frame within the window. This baseline model therefore shares the same model class (Transformer) as the seq2seq model but removes sequence generation on the output side.

\textbf{Seq2seq.} To evaluate the seq2seq architecture independently of cross-subject parameterization, we trained a model with the autoregressive encoder–decoder architecture described in Methods \ref{Methods: Model Architecture}, separately for each subject, with no cross-subject parameter sharing.

\subsubsection{Hyperparameter selection for temporal context and prediction target}
\label{Methods: Temporal model selection}

Each baseline model has temporal hyperparameters that affect its performance. This includes the stimulus context length \(k\), which determines how many TRs of stimulus history the model receives as input, and the prediction target, which specifies the output timepoint or subsequence to be predicted. Both temporal hyperparameters were selected separately for each model using a two-stage validation procedure.

We evaluated each configuration using the Pearson correlation between predicted and observed BOLD signals across 1,000 cortical parcels. For each parcel and subject, correlations were Fisher-transformed, then averaged across parcels and subjects to obtain a single validation metric \(\bar z\), which served as the objective for both stages.

\textbf{Step 1: Selecting the stimulus context length \(k\)}

For single-point prediction with the ridge and seq2one models, we performed a grid search over \(k\in\{5,10,\dots,50\}\). The input comprised the most recent \(k\) stimulus frames \((x_{t-k+1},\dots,x_t)\), the output was the current BOLD response \(y_t\). By including stimulus information from multiple preceding time points, the models could capture temporal dependencies associated with the delayed hemodynamic response. The stride of the sliding window was set to 1~TR. For sequence prediction with the seq2seq model, we searched \(k\in\{10,15,\ldots,50\}\) using the same stimulus windows, with the output defined as the contiguous sequence \((y_{t-k+6},\ldots,y_t)\). The first five time points of each window were excluded from the prediction target to provide sufficient preceding stimulus context, and a 5~TR sliding stride was used to reduce redundancy among overlapping windows.

\textbf{Step 2: Selecting the prediction target within fixed \(k\)}

After fixing each model’s selected \(k\), we selected the prediction target on the validation set to maximize \(\bar z\). For the ridge and seq2one models, we searched over the target index \(\tau\in\{5,10,\dots,k\}\), predicting \(y_{t-k+\tau}\)  with a 1 TR sliding stride. For the seq2seq model, we evaluated the start  \(a\) and end \(b\) of the output sequence in sequential sweeps, with the earliest output at least five TRs after window onset and a 5 TR stride. The hyperparameters \((k^{\ast},\tau^{\ast})\) or \((k^{\ast},[a^{\ast},b^{\ast}])\) were fixed for all subsequent test analyses.

\subsubsection{Baseline models for cross-subject parameterization}

We assessed the contribution of cross-subject parameterization by training the seq2seq models under three schemes that differ only in how subject information is used and how parameters are shared across participants.

\textbf{Individual-specific models.} We trained one model per subject using only that subject's data, resulting in fully independent subject-specific models. For this comparison, checkpoints from the same training epoch were used across participants.

\textbf{Pooled model.} We trained a single model on data from all subjects, using the same parameters to predict responses for every subject. No subject-specific components were included.

\textbf{Hybrid model.} As described in the Model Architecture section, the hybrid model shares encoder and decoder parameters across subjects, but includes a learnable subject embedding and subject-specific output projections. This allows shared representations of stimulus dynamics while capturing stable between-subject differences in response patterns.

\subsection{Adapting to new subjects with limited data}
\label{Methods:Cross subject transfer under data scarcity}

We evaluated whether we can leverage the model's ability to share representations across subjects to improve prediction in situations where subject-specific data are limited. In each analysis, one participant was designated as the target participant, and the remaining participants provided cross-subject information. We repeated the procedure with each participant serving as the target in turn and averaged performance across iterations. We considered two settings that differed in how cross-subject information was incorporated and in the availability of auxiliary data.

\subsubsection{Few-shot adaptation to new participants}
\label{Few-shot adaptation to new participants}

For each target participant, we first pretrained the hybrid model using data from the other three participants, including Friends, Bourne, Figures, and Life. We then froze the shared model parameters and fine-tuned only the subject-specific parameters using varying amounts of the target participant's Bourne data (12, 24, 36, 48, 60, 72, 84, and 96 minutes). Training subsets were nested, with successive Bourne runs added cumulatively as the amount of target-subject fine-tuning data increased. During this adaptation stage, only data from the target participant were used. For comparison, we trained an individual-specific model from scratch using the same amounts of target-participant Bourne data. Both models were evaluated on held-out Bourne data, which were excluded from pretraining and fine-tuning for all participants, to assess in-distribution generalization and on Wolf to assess out-of-distribution generalization.

As increasingly large multi-subject datasets become available, this setting provides a practical way to benefit from these accumulated data without repeatedly obtaining, storing, and retraining on the underlying fMRI datasets. Instead, researchers can reuse progressively improved pretrained models and adapt them directly to newly collected data from their own participants.

\subsubsection{Cross-subject learning across data-availability and distributional settings}
\label{Cross-subject learning across data-availability and distributional settings}

We next examined how the benefits of cross-subject learning vary with data availability and stimulus differences across three practical transfer scenarios. Under each transfer scenario, we compared an individual-specific model trained only on the target participant's limited data, a fully shared model trained across participants, and a hybrid cross-subject model that additionally incorporated data from the remaining participants.

\textbf{Case 1: Leveraging a large external dataset to improve prediction with limited data (large-out condition)}

This scenario represents a setting in which a researcher has collected a small amount of movie-viewing fMRI data from a new participant and seeks to improve prediction using a large public dataset containing different participants and stimuli. The target participant's \textit{Bourne} runs served as the limited subject-specific training data. The external training set comprised the other participants' \textit{Friends} data and their Movie10 runs from \textit{Bourne}, \textit{Figures}, and \textit{Life}. Model performance was evaluated on the target participant’s held-out \textit{Wolf} runs. \textit{Wolf} runs were held out from all  training data, so this assessed generalization to stimuli that were unseen during model fitting.

\textbf{Case 2: Leveraging data from other participants in the same dataset to improve within-stimulus prediction (small-in condition)}

This scenario represents a setting in which all participants viewed the same naturalistic stimulus, but only a limited number of runs are available from the target participant. We treated the runs from each Movie10 film as a separate small dataset and used run-level cross-validation. On each fold, one run from the target participant was held out for evaluation, and the remaining runs from that participant and film served as the limited subject-specific training data. The cross-subject model additionally incorporated runs from the other participants who viewed the same film. This case therefore tested how well the models leveraged limited data from other participants to improve prediction for held-out portions of the same stimulus.

\textbf{Case 3: Leveraging a small out-of-distribution dataset to improve prediction with limited data (small-out condition)}

This scenario represents a setting in which a researcher has collected a small movie-viewing fMRI dataset and has access to a comparably sized dataset in which different participants viewed a different stimulus. The target participant's \textit{Bourne} runs served as the limited subject-specific training data, while the other participants' \textit{Figures} runs served as the external training data. Model performance was evaluated on the target participant’s \textit{Wolf} runs, which were excluded from all training data. In contrast to Case 1, this case tested how well the models leveraged a relatively small external dataset to improve generalization to a novel stimulus.

\subsection{AI Transparency and Reproducibility Statement}

This study used pretrained neural network models, including VideoMAE, HuBERT, Qwen, and BridgeTower, to derive multimodal representations of the audiovisual stimuli. The specific models, feature extraction procedures, selected model layers, temporal aggregation procedures, and representation dimensions are described in Section 2.2. The architecture, training objective, optimization procedure, data splits, and evaluation procedures for the proposed Transformer-based brain encoding model are described in the corresponding Methods sections. Hyperparameter selection was performed using validation data only, and the selected hyperparameters were fixed before evaluation on the held-out test data.

Code used for feature extraction, model training, and evaluation is available at \href{https://github.com/Angelneer926/multimodal-seq2seq-brain-encoding/tree/main}{GitHub repository}.

During the preparation of this work, the authors used ChatGPT (OpenAI) to improve the language, grammar, and readability of the manuscript and to assist with code refinement and debugging. The tool was not used to independently conduct analyses or draw scientific conclusions. The authors reviewed and edited all AI-assisted content as needed and take full responsibility for the content of the publication.

\section{Results}

We evaluated our models on the subset of Courtois NeuroMod (CNeuroMod) naturalistic-viewing corpus comprising four deeply sampled participants (sub-01/02/03/05). Participants underwent fMRI while watching six seasons of \textit{Friends} and four feature-length films, including \textit{Bourne}, \textit{Figures}, \textit{Life}, and \textit{Wolf}. Unless otherwise noted, we used the fixed stimulus-wise split described in Section~\ref{Methods splits}: \textit{Friends} Seasons 1--5 for training, Season 6 Episodes 1--12 for validation, Season 6 Episodes 13--24 for in-distribution testing, and the four films for out-of-distribution evaluation.

\subsection{Sequence modeling improves within-subject prediction of brain activity from naturalistic audiovisual input}

To assess the benefits of sequence modeling for predicting brain activity, we compared three approaches that differ in how they map stimulus information to neural responses: (i) ridge regression, which predicts a single neural time point from a temporal window of preceding stimulus features and represents the most widely used approach in previous studies; (ii) transformer seq2one, which also predicts a single neural time point from a temporal window of preceding stimulus features, but uses a transformer encoder to model temporal dependencies across the input sequence; and (iii) transformer seq2seq, which predicts a sequence of neural responses from a sequence of stimulus features, allowing the model to capture temporal dependencies across both stimulus and neural time series. We were first interested in evaluating how model architecture influences predictive performance within participants, so each model was trained and tested separately for every subject.

We selected temporal hyperparameters on the validation split to ensure a fair comparison across modeling approaches (see Methods \ref{Methods: Temporal model selection}). Guided by prior evidence that sensory cortices integrate over short windows of a few seconds whereas higher-order regions exhibit temporal receptive windows spanning tens of seconds to minutes \citep{hasson2015hierarchical}, we performed a grid search over stimulus context length \(k\) over a range that captures short-to-intermediate timescales while remaining computationally tractable for sequence models. Validation performance varied systematically with stimulus context length across the three models.  For ridge regression, performance was highest at a relatively short context length of  \(k=15\) TRs  whereas transformer seq2one achieved its highest performance at \(k=30\) TRs. Transformer seq2seq continued to improve as the context length increased before reaching a plateau at longer windows, with optimal \(k=45\) TRs. Validation curves are shown in Supplementary Materials (Fig.~\ref{sup_1}A).

After selecting the optimal \(k\) for each model, we performed a second grid search to optimize the prediction target. For the ridge regression and transformer seq2one models, the predicted fMRI time point was varied in 5-TR increments from TR~5 to TR~\(k\) with a 1-TR sliding window. Validation performance was highest at target frames 15 and 25 for the ridge regression and seq2one models, respectively (Fig.~\ref{sup_1}B--C). For the  transformer seq2seq model, the best-performing contiguous output segment was TRs~11--40 (Fig.~\ref{sup_1}D--E). These target timeframes were used in all subsequent analyses.

We evaluated all models on the held-out test set comprising the in-distribution \textit{Friends} split and the out-of-distribution \textit{Movies} set. Average model performance was highest for the seq2seq model (mean $\pm$ s.e.m; \textit{Friends}: $r = 0.293\pm0.016$; \textit{Movies}: $r = 0.201\pm0.011$), followed by the seq2one model (\textit{Friends}: $r = 0.278\pm0.016$; \textit{Movies}: $r = 0.188\pm0.011$), then the ridge regression model (\textit{Friends}: $r = 0.250\pm0.015$; \textit{Movies}: $r = 0.151\pm0.010$). Participant-level results are reported in Supplementary Table~\ref{tab:s1}. Given the small sample size ($n=4$), we did not perform group-level inferential tests. Instead, we emphasize effect sizes and cross-participant consistency, with all four participants showing a consistent pattern of model performance.

\begin{figure}[htbp]
\begin{center}
\includegraphics[
    page=1,
    width=1\textwidth,
    trim=0cm 3cm 0cm 0cm,
    clip
]{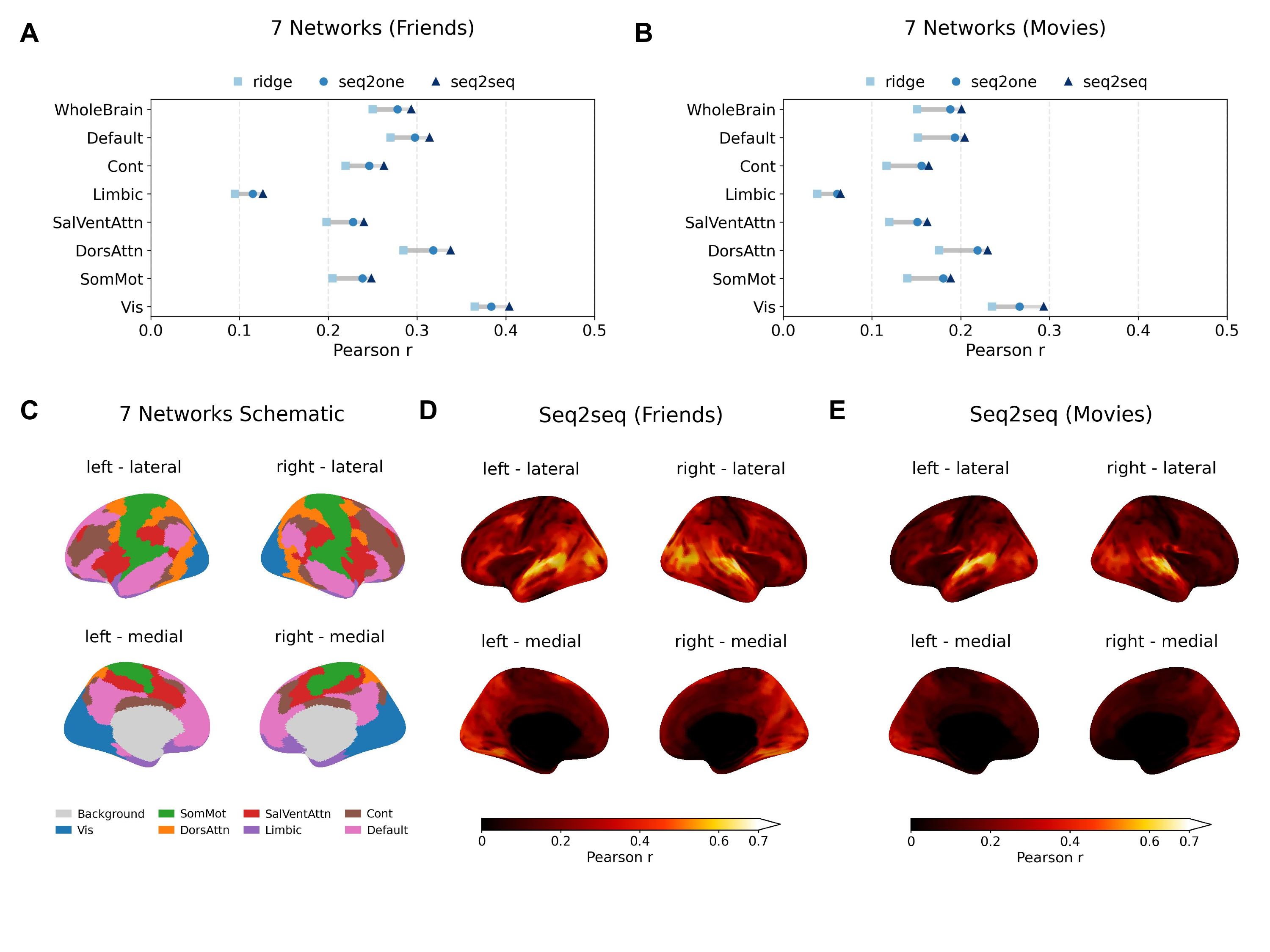}
\caption{\textbf{Performance of ridge, seq2one, and seq2seq across cortical networks and whole brain.} \textbf{(A–B)} Mean Pearson correlations \(r\) for ridge regression, seq2one, and seq2seq on Friends (A) and Movies (B), averaged across subjects. \textbf{(C)} Cortical parcellation into seven canonical networks used for network-level analyses. \textbf{(D–E)} Whole-brain prediction maps for seq2seq on Friends and Movies. Network abbreviations: Vis, visual; SomMot, somatomotor; DorsAttn, dorsal attention; SalVentAttn, salience/ventral attention; Limbic, limbic; Cont, control; Default, default mode.}
\label{results_part1}
\end{center}
\end{figure}

The same ordering held across all seven networks in both datasets, with the seq2seq model consistently outperforming ridge regression and showing additional gains over the seq2one model (Fig.~\ref{results_part1}A-B). For \textit{Friends}, the improvement of seq2seq over ridge ranged from \(\Delta r = 0.032 \) in the Limbic network to  \(\Delta r = 0.053 \) in the Dorsal Attention network. Relative to the seq2one model,  the largest improvements were observed in the Visual (\(\Delta r = 0.020\)) and Dorsal Attention networks (\(\Delta r = 0.020\)) followed by the Control network (\(\Delta r = 0.016\)). For \textit{Movies}, the gains over ridge were larger overall, ranging from  (\(\Delta r = 0.026\)) in the Limbic network to (\(\Delta r = 0.058\)) in the Visual network. The largest improvement over seq2one was also observed in the Visual network (\(\Delta r = 0.027\)) with additional gains in the Dorsal Attention  network (\(\Delta r = 0.011\)) and Salience/Ventral Attention (\(\Delta r = 0.011\)). Together, these results indicate that explicit sequence modeling improves prediction across cortical networks, with larger gains under out-of-distribution evaluation.

Parcel-wise analyses showed a similar spatial pattern. Seq2one improved over ridge across sensory and association cortices, while seq2seq produced additional gains across prefrontal, parietal, and temporal regions for both \textit{Friends} and \textit{Movies} (Fig. ~\ref{sup_2}A-D). The seq2seq prediction maps further showed predictive performance extending from early sensory cortex into higher-order control and default-mode regions (Fig.\ref{results_part1}D–E), indicating that the benefits of sequence modeling were broadly distributed across the cortex. 

\subsection{Hybrid parameterization is optimal for cross-subject prediction}
\label{results:2}
We next investigated how different parameter-sharing strategies affect cross-subject modeling performance. Specifically, we compared three approaches: a fully shared model, where all parameters are shared across participants; a fully individual model, where a separate model is trained for each subject; and a hybrid model, which combines a shared Transformer encoder with a partial subject-specific decoder consisting of individualized embeddings and linear output layers. All models used the Transformer seq2seq architecture, which performed best in the preceding comparison of sequence-modeling approaches.

\begin{figure}[htbp]
\begin{center}
\includegraphics[
    page=2,
    width=1\textwidth,
    trim=0cm 3cm 0cm 0cm,
    clip
]{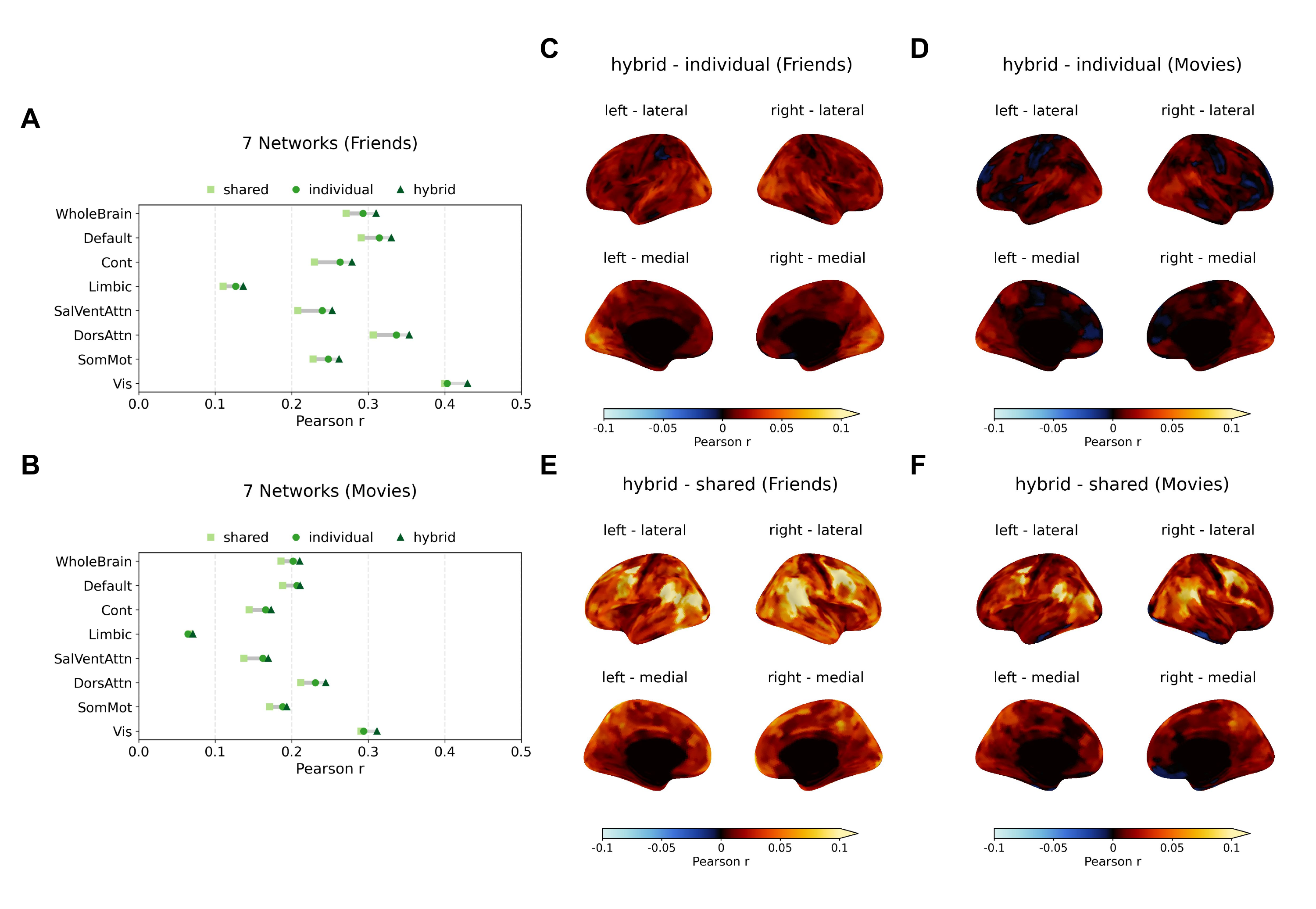}
\caption{\textbf{Performance of fully shared, fully individual, and hybrid models across cortical networks and whole brain.} \textbf{(A–B)} Mean Pearson correlations \(r\) for fully shared, fully individual, and hybrid (shared encoder + partial subject-specific decoder) models on Friends (A) and Movies (B), averaged across subjects.
\textbf{(C–D)} Difference maps (hybrid $-$ individual) show advantages in early sensory and attention-related cortices, including visual, somatomotor, and dorsal attention regions. \textbf{(E–F)} Difference maps (hybrid $-$ shared) highlight advantages in higher-order association networks, including default-mode, control, and temporoparietal regions.}
\label{results_part2}
\end{center}
\end{figure}

Across both datasets, the fully individual model outperformed the fully shared model at the whole-brain level. For \textit{Friends}, prediction accuracy increased from r = 0.271 for the shared model to r = 0.293 for the individual model (\(\Delta\) r = 0.022). For \textit{Movies}, performance increased from r = 0.186 to r = 0.202 (\(\Delta\) r = 0.016; Fig.~\ref{results_part2}A--B). This advantage indicates that retaining subject-specific parameters improves prediction relative to fully pooling participants. The hybrid model further outperformed the individual model in both datasets, reaching r = 0.310 for \textit{Friends} (\(\Delta\) r = 0.017 relative to individual and \(\Delta\) r = 0.039 relative to shared), and r = 0.210 for \textit{Movies} (\(\Delta\) r = 0.008 relative to individual and \(\Delta\) r = 0.024 relative to shared). The hybrid model also achieved the highest performance across all seven cortical networks in both datasets, suggesting that shared and subject-specific components provide complementary predictive benefits. Whole-brain prediction maps for the hybrid model on \textit{Friends} and \textit{Movies} are shown in Supplementary Fig.~\ref{sup_3}.


Spatial analyses revealed distinct patterns for the hybrid--individual and hybrid--shared contrasts (Fig.~\ref{results_part2}C--F). Relative to the individual model, the largest network-level improvements for the hybrid model were observed in the Visual (\(\Delta\) r = 0.027) and Dorsal Attention (\(\Delta\) r = 0.017) networks for \textit{Friends}. The same networks showed the largest gains for \textit{Movies} (\(\Delta\) r = 0.017 and 0.014, respectively), consistent with prior evidence that these systems exhibit reliable stimulus-driven responses during naturalistic viewing \citep{ren2017inter}. Parcel-wise maps provided a finer-grained view of the spatial distribution of these effects (Fig.~\ref{results_part2}C--D). 

The hybrid--shared comparison showed a broader pattern of improvement, extending to higher-order association networks. Particularly strong gains were observed in the Control and Default networks, reaching \(\Delta\) r = 0.049 and 0.040, respectively, for \textit{Friends}, with corresponding gains of \(\Delta\) r = 0.029 and 0.023 for \textit{Movies} (Fig.~\ref{results_part2}E--F). These higher-order networks are known to exhibit greater inter-individual variability, consistent with their involvement in internally oriented cognition and complex integrative processes that can differ across individuals \citep{vanderwal2017individual,yeshurun2021default}.

Together, these results suggest that hybrid parameterization leverages the strengths of both approaches: it captures shared representational structure where subjects’ neural dynamics are aligned, while preserving subject-specific detail in regions with high inter-individual variability. As a result, this approach achieves more robust and generalizable prediction across the cortex compared to either fully shared or fully individual models. As in the previous section, results are summarized at the group-averaged level, with per-subject performance values reported in Supplementary Table~\ref{tab:s2}.

\subsection{Rapid adaptation with limited individual data}

We next investigated whether an existing model pretrained on other participants could be rapidly personalized to a new individual using only limited target-subject fMRI data (see Methods \ref{Few-shot adaptation to new participants}). To this end, we compared two approaches: a fully individual model trained from scratch and a hybrid model initialized from a model pretrained on all other subjects and subsequently fine-tuned only on subject-specific parameters. Both models used the same amounts of target-participant Bourne data (12, 24, 36, 48, 60, 72, 84, and 96 minutes) for subject-specific training or adaptation. Performance was evaluated on held-out Bourne data (ID) and on Wolf (OOD). All models used the Transformer seq2seq hybrid backbone identified in the previous section and were trained under identical optimization settings, enabling a direct test of whether multi-subject pretraining improves the data efficiency and stability of subject-specific adaptation under realistic constraints.

\begin{figure}[htbp]
\begin{center}
\includegraphics[page=3,width=1\textwidth]{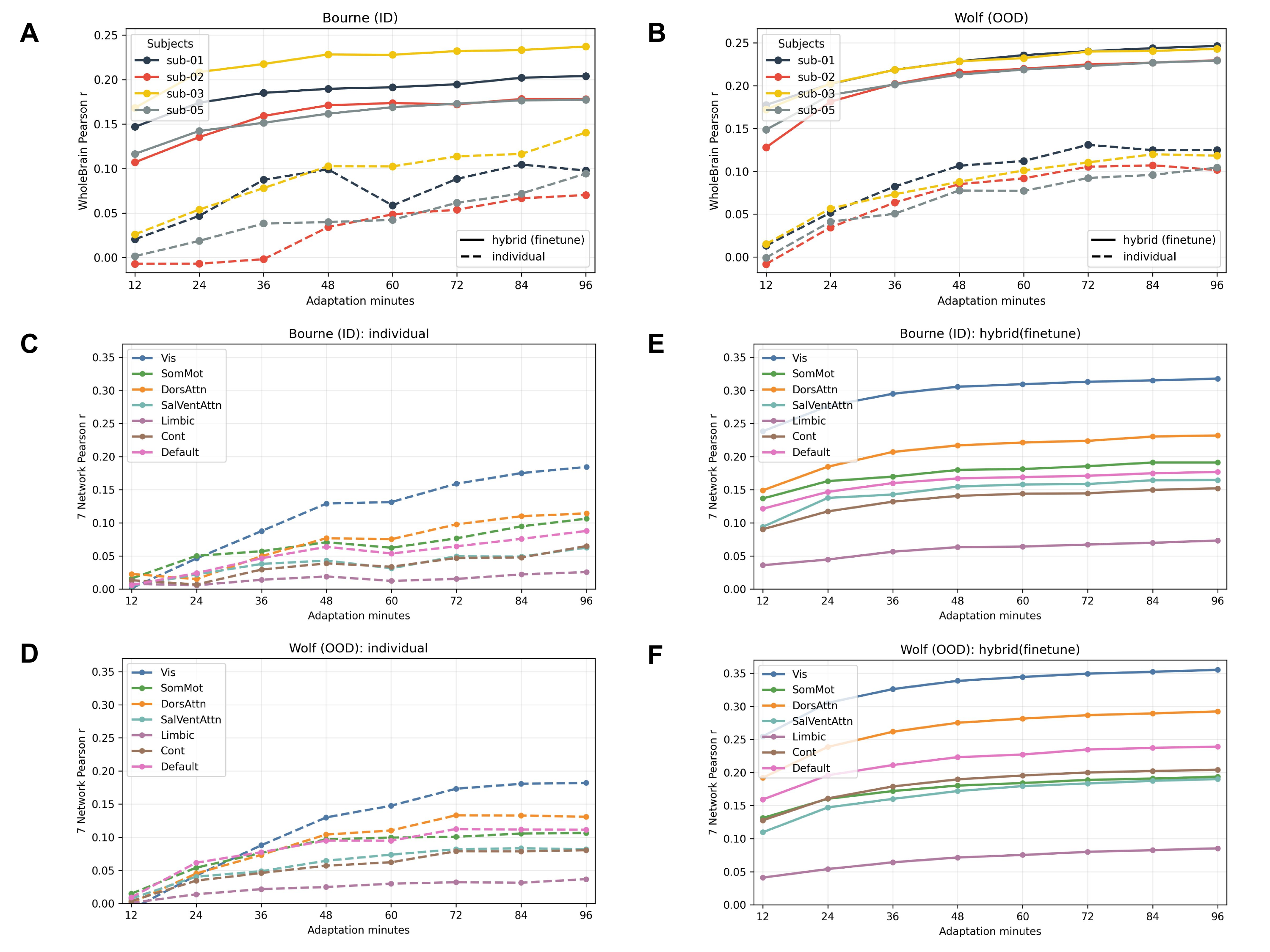}
\caption{\textbf{Data-efficient adaptation: comparison of individual and hybrid (fine-tuned) models across data scales and cortical networks.} \textbf{(A–B)} Whole-brain learning curves showing prediction performance as a function of available target-subject training data. Models were trained or fine-tuned on Bourne (12–96 min) and evaluated on either the same movie (Bourne, in-distribution; A) or a different movie (Wolf, out-of-distribution; B). Solid lines denote hybrid models fine-tuned from multi-subject pretraining, while dashed lines denote individual models trained from scratch. \textbf{(C–D)} Network-level learning curves for individual model evaluated on Bourne (C) and Wolf (D), showing modest gains and high variability across networks. \textbf{(E–F)} Network-level learning curves for hybrid fine-tuned model evaluated on Bourne (E) and Wolf (F), revealing consistently higher accuracy and clear, monotonic improvements across all networks.}
\label{results_part3}
\end{center}
\end{figure}

Learning curves revealed that subject-specific adaptation occurred rapidly, with the fine-tuned hybrid model outperforming the individual model trained from scratch even when only 12 minutes of data were available (subject-averaged gain: Bourne +0.124; Wolf +0.151; Fig.\ref{results_part3} A-B). Remarkably, this small amount of fine-tuning data was sufficient to exceed the performance of an individual model trained with 96 minutes, indicating that shared representations acquired from other participants provide a powerful inductive prior that enables efficient personalization. In contrast, models trained from scratch on the same small datasets achieved lower accuracy and showed less consistent gains across training-set sizes. As the amount of training data increased, the fine-tuned hybrid model improved monotonically in both in-distribution and out-of-distribution evaluations, with diminishing marginal gains at larger data scales. While performance continued to increase through 96 minutes, most improvements were achieved within roughly the first hour, suggesting that the proposed fine-tuning procedure offers a practical and data-efficient route to individualized brain encoding.

A network-level analysis of adaptation dynamics revealed that, across datasets, both modeling approaches followed a broadly similar hierarchical ordering, in which sensory systems, especially the visual network, showed the fastest and most pronounced gains with limited data, whereas higher-order association networks generally improved more slowly with increasing data (Fig. \ref{results_part3}C–F). Against this shared hierarchical profile, the two approaches diverged in both the starting point and the reliability of network-level learning curves. Individually trained models (Fig.\ref{results_part3}C–D) displayed shallow and often irregular improvements, with several association networks plateauing prematurely or exhibiting fluctuations as data increased. Although the visual system improved steadily, most higher-order regions remained difficult to predict even at the largest data scale. In contrast, the hybrid fine-tuned model (Fig.\ref{results_part3}E–F) demonstrated smoother, more monotonic learning trajectories across nearly all networks, achieving higher peak performance and sustained improvement well beyond the saturation point of individual models.

\subsection{Transfer performance across practical data-scarcity scenarios}

\begin{figure}[htbp]
\begin{center}
\includegraphics[width=0.6\textwidth]{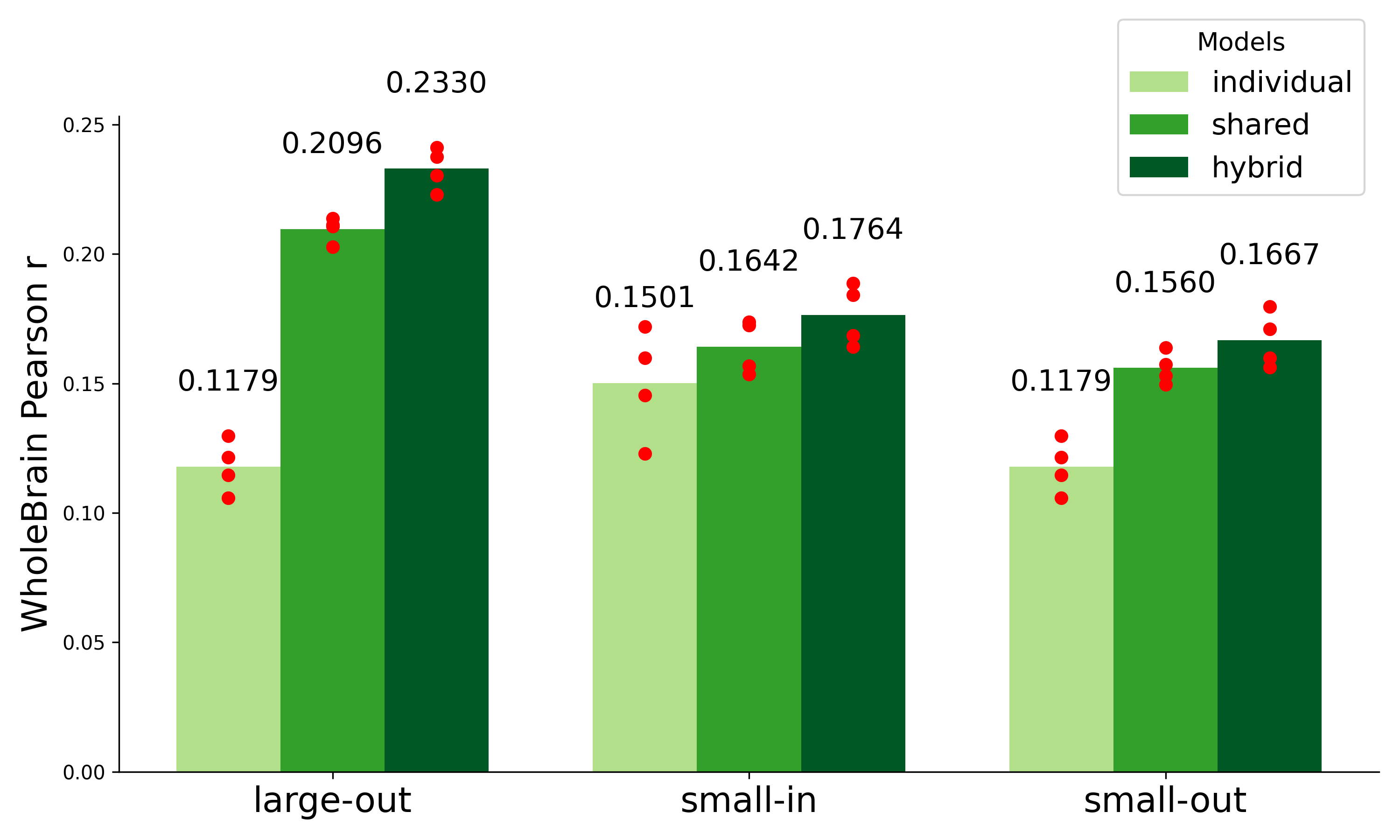}
\caption{\textbf{Transfer performance of individual, shared, and hybrid models under three representative data-scarcity scenarios.} Bars show the mean whole-brain Pearson correlation across subjects, and red dots indicate subject-wise performance (one dot per subject). The hybrid model consistently achieved the highest whole-brain prediction accuracy across all transfer scenarios.}
\label{results_part4}
\end{center}
\end{figure}

We further examined how the three modeling approaches (individual, shared, hybrid) performed under three representative transfer scenarios (see Methods \ref{Cross-subject learning across data-availability and distributional settings}), designed to capture differences in both data scale and distribution (Fig.\ref{results_part4}). The advantage of hybrid models was most pronounced in the large-out condition, where abundant data from other individuals were available but the target subject had only limited samples. In this setting, both the shared and hybrid models substantially outperformed the individual baseline (0.2096 and 0.2330 vs. 0.1179), with the hybrid model achieving the best overall performance. In contrast, the performance gap among models was smallest in the small-in scenario (hybrid: 0.1764; shared: 0.1642; individual: 0.1501), where both the target and auxiliary datasets were limited and the test data were drawn from the same distribution as training. For completeness, the small-in evaluation is repeated across all films, and the detailed film-wise results are reported in Supplementary Table~\ref{tab:s3}. Here, the strong overlap between training and test stimuli reduces the relative advantage of transfer-based approaches: although the hybrid model still achieved the best performance, the limited diversity and scale of the auxiliary data constrained the potential benefits of cross-subject representation sharing. The small-out condition, where all data were scarce but generalization was evaluated on a new stimulus distribution, showed an intermediate pattern (hybrid: 0.1667; shared: 0.1560; individual: 0.1179). Even under this challenging condition, transfer-based models provided meaningful advantages over the individual baseline, indicating that cross-subject information can still introduce useful inductive biases for adaptation to unseen stimuli.

Taken together, these results reveal a systematic dependency of model performance on the amount of subject-specific data available: when the target dataset is small, transfer-based approaches outperform individually trained models, reflecting their ability to compensate for data scarcity by exploiting shared representational structure. When more subject-specific data are available, as shown in Section \ref{results:2}, the performance gap narrows and individually trained models can surpass a model trained on multiple participants without a hybrid architecture. Across all conditions, however, the hybrid model consistently achieved the best performance, consistent with its ability to integrate population-level priors with individual-specific adaptation across diverse data regimes. 

\section{Discussion}

Naturalistic brain encoding requires models that can capture the temporal structure of continuous experience, integrate information across stimulus modalities, and generalize across individuals. In this study, we developed a unified framework that brings these components together within a single model and training objective. We evaluated this approach under both in-distribution (held-out \textit{Friends}) and out-of-distribution (held-out films) conditions. Across participants and large-scale networks, several consistent findings emerged. Explicitly modeling brain responses as a sequence with the sequence-to-sequence Transformer (seq2seq) yielded consistent gains over both linear ridge and single-timepoint Transformer (seq2one) predictors, with larger improvements under stimulus distribution shift. A hybrid cross-subject parameterization consistently outperformed both fully shared and fully individual models, highlighting the complementary benefits of shared stimulus representations and subject-specific neural readouts. This advantage extended to data-scarce settings, where cross-subject pretraining substantially improved the efficiency of personalization. Together, these findings suggest that jointly modeling temporal dynamics and cross-subject information can improve both the accuracy and data efficiency of naturalistic brain encoding.

The advantage of sequence modeling likely reflects the temporal structure inherent in both naturalistic experience and measured brain responses. Prior encoding studies have shown that incorporating preceding stimulus context can improve the prediction of brain responses \citep{jain2018incorporating}, while autoregressive modeling has demonstrated that past BOLD responses can also predict subsequent neural activity \citep{paugam2024benchmark}. Our seq2seq model combines these two forms of temporal information by conditioning predictions on both the evolving multimodal stimulus context and preceding neural responses. However, the present analyses do not isolate the extent to which the benefit of response history reflects stimulus-dependent neural dynamics versus intrinsic and hemodynamic temporal structure.

Beyond the benefit of response history itself, the context-length analyses suggest that different models vary in how effectively they use information over extended temporal windows. Previous studies have shown that encoding models can benefit from incorporating information over longer or multiple temporal scales \citep{jain2020interpretable,jindal2026does}. Here, we further find that these benefits depend on model architecture. For ridge regression, increasing context length introduces more correlated and potentially redundant lagged predictors, which may limit the marginal benefit of additional stimulus information despite regularization. Seq2one provides greater flexibility by using attention to weight temporal information according to the current input, allowing the model to emphasize more informative portions of a longer context. Seq2seq further benefits from sequence-level supervision, as each temporal context is optimized against multiple adjacent brain-response targets rather than a single time-point target. Information that remains predictive across neighboring responses can therefore be reinforced across multiple outputs. This may support more stable use of longer temporal contexts and help explain the more sustained gains observed with increasing context length.

The advantage of the hybrid cross-subject strategy likely arises from separating components that can be shared across individuals from those that  remain subject-specific. Previous work has highlighted the importance of preserving individual-specific information in brain models \citep{paugam2024benchmark,freteault2025alignment}. This subject specificity is consistent with substantial inter-individual variability at multiple levels, including variability in hemodynamic timing and response amplitude \citep{handwerker2004variation}, residual differences in functional topographies, and individual variation in higher-order cognitive responses to complex naturalistic stimuli \citep{mueller2013individual}. Our results further suggest that preserving individual-specific information does not require the entire encoding model to be estimated separately for each participant. The shared encoder can pool data across participants to learn broadly transferable stimulus representations, capturing common multimodal structure in the narratives and stimulus–response relationships that are shared across individuals. In contrast, subject-specific parameters within the decoder allow the model to accommodate stable differences in individual neural responses that persist even after anatomical normalization. Accordingly, this division likely underlies the consistent advantage of the hybrid architecture over both fully shared and fully individual models across datasets.

The rapid adaptation results further highlight the value of hybrid parameterization when subject-specific data are limited. By retaining a shared representational scaffold learned from other participants and updating only lightweight subject-specific components, the model can leverage population-level information while efficiently adapting to a new individual. Previous work has similarly shown that information learned from other participants can substantially reduce the amount of target-subject data required to construct personalized encoding models \citep{wen2018transferring,gu2022personalized}. Notably, with only 12 minutes of target-subject data for fine-tuning, the pretrained hybrid model outperformed an individual model trained from scratch with 96 minutes of data, demonstrating a substantial gain in personalization efficiency. This finding shares a central motivation with recent cross-participant mapping work \citep{tang2026rapid}, although the approaches differ in how cross-subject information is incorporated. Whereas their approach transfers reference encoding models through functional alignment using shared stimuli, ours incorporates cross-subject parameter sharing directly into the encoding architecture. A practical implication of our approach is that it can integrate datasets collected with different stimuli without requiring a separate shared alignment dataset. This flexibility is demonstrated in our small-out condition, where the hybrid model improved prediction using non-overlapping training stimuli from target and auxiliary participants. Meanwhile, consistent with their findings, our transfer analyses showed that the benefits of cross-subject learning depend on the relative availability of target-subject and auxiliary data. The largest gains emerged when target-subject data were limited but substantial auxiliary data were available, whereas the gains were smaller when both sources of data were scarce.

Network-level analyses provided a more detailed view of how shared and subject-specific information contribute across cortical systems. The benefits of sharing were most pronounced in sensory and attention networks, where responses are strongly constrained by common perceptual input and externally driven attention \citep{ren2017inter}. Given the inherently noisy nature of fMRI, pooling data across participants may further improve estimation of reliable stimulus-related response structure in these systems. In contrast, subject-specific modeling contributed more strongly in higher-order association networks, where responses may depend more on individual differences in prior knowledge, memory, interpretation, and internal context \citep{vanderwal2017individual,yeshurun2021default}. Complete parameter sharing may therefore obscure meaningful individual response patterns in these regions.

This network dependence was also evident in the learning curves for personalization. Sensory networks may be comparatively straightforward to calibrate from limited data, whereas association cortex may benefit from a better-initialized representational scaffold that supports stable incremental updates. This observation suggests a network-aware view of personalization, in which subject-specific capacity, regularization, or fine-tuning emphasis could be weighted toward higher-order networks to better capture individual-specific components of long-timescale, integrative processing. 
 
 Several limitations should be considered. First, although each participant contributed extensive naturalistic fMRI data, the small number of participants (\(N=4\)) limits the extent to which our findings can be generalized at the population level. The consistency of the main effects across all four participants provides some reassurance, but larger samples will be needed to establish the robustness of these effects across individuals. Second, the hybrid architecture evaluated here represents only one configuration within a broader design space. We did not systematically test which modules or parameters of the model should be shared across individuals and which should remain subject-specific. Future ablation studies could compare alternative designs to identify strategies that further improve cross-subject prediction and personalization. Finally, the current approach used fixed-length context windows. Although this provides a tractable way to model temporal dependencies, the resulting boundaries may not align with event structure in the stimulus or transitions in latent brain states. Given evidence that event boundaries provide meaningful units for segmenting and organizing continuous experience \citep{zacks2007event,baldassano2017discovering}, an important direction is to incorporate event-aware segmentation into future models.

 More broadly, our findings extend prior naturalistic encoding approaches by showing how temporal sequence modeling and cross-subject learning can be leveraged to improve both prediction and personalization. These advances have several practical implications. In fMRI studies, extensive data collection from each individual is often constrained by scanning time and cost. As larger public naturalistic neuroimaging datasets become available, cross-subject pretraining could allow these data to serve as increasingly strong population priors for adapting encoding models to new individuals. This could improve data efficiency and lower the data requirements for individualized encoding. Beyond personalization, stronger and more transferable encoding models may also provide a useful foundation for downstream applications, including neural decoding, reconstruction, and cross-task generalization. Looking further ahead, a particularly promising direction is the development of memory-augmented encoding models. Recent modeling work suggests that information from preceding narrative events can be selectively accessed when it becomes relevant to ongoing processing \citep{song2025neural}. This motivates memory-augmented encoding models that incorporate temporally distant information through selective retrieval rather than uniformly extending the context window. This could also serve as a computational framework for testing hypotheses about when human memory retrieval occurs, what past information is selected, and how retrieved information shapes ongoing neural responses. Taken together, these directions may support more scalable and long-timescale models of brain responses during naturalistic experience.

\section*{Data and Code Availability}

The fMRI data and stimulus materials analyzed in this study are publicly available through the Algonauts Project 2025 challenge, based on the Courtois NeuroMod dataset (\url{https://algonautsproject.com/2025/challenge.html}). All code used in the analysis is publicly available at \url{https://github.com/Angelneer926/multimodal-seq2seq-brain-encoding}.

\section*{Author Contributions}

Q.H.: Conceptualization, methodology, software, formal analysis, investigation, data curation, validation, visualization, writing—original draft, writing—review $\&$ editing.
M.D.R.: Conceptualization, methodology, supervision, writing—review $\&$ editing.
Y.C.L.: Conceptualization, methodology, supervision, project administration, writing—review $\&$ editing.

\section*{Ethics}
All subjects provided informed consent to participate in this study, which was approved by the ethics review board of the “CIUSSS du centre-sud-de-l’île-de-Montréal” (under number CER VN 18-19-22).

\section*{Declaration of Competing Interests}

The authors declare no competing interests.

\section*{Acknowledgments}

The Courtois project on neural modelling was made possible by a generous donation from the Courtois foundation, administered by the Fondation Institut Gériatrie Montréal at CIUSSS du Centre-Sud-de-l’île-de-Montréal and University of Montreal. The Courtois NeuroMod team is based at Centre de Recherche de l’Institut Universitaire de Gériatrie de Montréal, with several other institutions involved. See the CNeuroMod documentation for an up-to-date list of contributors (\url{https://docs.cneuromod.ca}).

Computing resources were provided by the University of Chicago Data Science Institute.

\printbibliography

@article{baldassano2017discovering,
  title={Discovering event structure in continuous narrative perception and memory},
  author={Baldassano, Christopher and Chen, Janice and Zadbood, Asieh and Pillow, Jonathan W and Hasson, Uri and Norman, Kenneth A},
  journal={Neuron},
  volume={95},
  number={3},
  pages={709--721},
  year={2017},
  publisher={Elsevier}
}

@article{hasson2015hierarchical,
  title={Hierarchical process memory: memory as an integral component of information processing},
  author={Hasson, Uri and Chen, Janice and Honey, Christopher J},
  journal={Trends in cognitive sciences},
  volume={19},
  number={6},
  pages={304--313},
  year={2015},
  publisher={Elsevier}
}

@inproceedings{boyle2023courtois,
  title={The Courtois NeuroMod project: quality assessment of the initial data release (2020)},
  author={Boyle, Julie and Pinsard, Basile and Borghesani, Valentina and Paugam, Francois and DuPre, Elizabeth and Bellec, Pierre},
  booktitle={2023 Conference on Cognitive Computational Neuroscience},
  pages={2023--1602},
  year={2023}
}

@article{naselaris2011encoding,
  title={Encoding and decoding in fMRI},
  author={Naselaris, Thomas and Kay, Kendrick N and Nishimoto, Shinji and Gallant, Jack L},
  journal={Neuroimage},
  volume={56},
  number={2},
  pages={400--410},
  year={2011},
  publisher={Elsevier}
}

@article{huth2016natural,
  title={Natural speech reveals the semantic maps that tile human cerebral cortex},
  author={Huth, Alexander G and De Heer, Wendy A and Griffiths, Thomas L and Theunissen, Fr{\'e}d{\'e}ric E and Gallant, Jack L},
  journal={Nature},
  volume={532},
  number={7600},
  pages={453--458},
  year={2016},
  publisher={Nature Publishing Group UK London}
}

@article{dupre2025voxelwise,
  title={The Voxelwise Encoding Model framework: a tutorial introduction to fitting encoding models to fMRI data},
  author={Dupr{\'e} la Tour, Tom and Visconti di Oleggio Castello, Matteo and Gallant, Jack L},
  journal={Imaging Neuroscience},
  volume={3},
  pages={imag\_a\_00575},
  year={2025},
  publisher={MIT Press 255 Main Street, 9th Floor, Cambridge, Massachusetts 02142, USA~…}
}

@article{frost2012measuring,
  title={Measuring structural--functional correspondence: spatial variability of specialised brain regions after macro-anatomical alignment},
  author={Frost, Martin A and Goebel, Rainer},
  journal={Neuroimage},
  volume={59},
  number={2},
  pages={1369--1381},
  year={2012},
  publisher={Elsevier}
}

@article{haxby2020hyperalignment,
  title={Hyperalignment: Modeling shared information encoded in idiosyncratic cortical topographies},
  author={Haxby, James V and Guntupalli, J Swaroop and Nastase, Samuel A and Feilong, Ma},
  journal={elife},
  volume={9},
  pages={e56601},
  year={2020},
  publisher={eLife Sciences Publications, Ltd}
}

@article{chen2015reduced,
  title={A reduced-dimension fMRI shared response model},
  author={Chen, Po-Hsuan Cameron and Chen, Janice and Yeshurun, Yaara and Hasson, Uri and Haxby, James and Ramadge, Peter J},
  journal={Advances in neural information processing systems},
  volume={28},
  year={2015}
}

@article{freteault2025alignment,
  title={Alignment of auditory artificial networks with massive individual fMRI brain data leads to generalisable improvements in brain encoding and downstream tasks},
  author={Freteault, Maelle and Le Clei, Maximilien and Tetrel, Loic and Bellec, Lune and Farrugia, Nicolas},
  journal={Imaging Neuroscience},
  volume={3},
  pages={imag\_a\_00525},
  year={2025},
  publisher={MIT Press 255 Main Street, 9th Floor, Cambridge, Massachusetts 02142, USA~…}
}

@article{wen2018transferring,
  title={Transferring and generalizing deep-learning-based neural encoding models across subjects},
  author={Wen, Haiguang and Shi, Junxing and Chen, Wei and Liu, Zhongming},
  journal={NeuroImage},
  volume={176},
  pages={152--163},
  year={2018},
  publisher={Elsevier}
}

@article{gu2022personalized,
  title={Personalized visual encoding model construction with small data},
  author={Gu, Zijin and Jamison, Keith and Sabuncu, Mert and Kuceyeski, Amy},
  journal={Communications Biology},
  volume={5},
  number={1},
  pages={1382},
  year={2022},
  publisher={Nature Publishing Group UK London}
}

@article{he2025multimodal,
  title={A Multimodal Seq2Seq Transformer for Predicting Brain Responses to Naturalistic Stimuli},
  author={He, Qianyi and Leong, Yuan Chang},
  journal={arXiv preprint arXiv:2507.18104v2},
  year={2025},
  url={https://arxiv.org/abs/2507.18104v2}
}

@article{gifford2024algonauts,
  title={The algonauts project 2025 challenge: How the human brain makes sense of multimodal movies},
  author={Gifford, Alessandro T and Bersch, Domenic and St-Laurent, Marie and Pinsard, Basile and Boyle, Julie and Bellec, Lune and Oliva, Aude and Roig, Gemma and Cichy, Radoslaw M},
  journal={arXiv preprint arXiv:2501.00504},
  year={2025}
}

@article{schaefer2018local,
  title={Local-global parcellation of the human cerebral cortex from intrinsic functional connectivity MRI},
  author={Schaefer, Alexander and Kong, Ru and Gordon, Evan M and Laumann, Timothy O and Zuo, Xi-Nian and Holmes, Avram J and Eickhoff, Simon B and Yeo, BT Thomas},
  journal={Cerebral cortex},
  volume={28},
  number={9},
  pages={3095--3114},
  year={2018},
  publisher={Oxford University Press}
}

@article{gorgolewski2016brain,
  title={The brain imaging data structure, a format for organizing and describing outputs of neuroimaging experiments},
  author={Gorgolewski, Krzysztof J and Auer, Tibor and Calhoun, Vince D and Craddock, R Cameron and Das, Samir and Duff, Eugene P and Flandin, Guillaume and Ghosh, Satrajit S and Glatard, Tristan and Halchenko, Yaroslav O and others},
  journal={Scientific data},
  volume={3},
  number={1},
  pages={1--9},
  year={2016},
  publisher={Nature Publishing Group}
}

@article{esteban2019fmriprep,
  title={fMRIPrep: a robust preprocessing pipeline for functional MRI},
  author={Esteban, Oscar and Markiewicz, Christopher J and Blair, Ross W and Moodie, Craig A and Isik, A Ilkay and Erramuzpe, Asier and Kent, James D and Goncalves, Mathias and DuPre, Elizabeth and Snyder, Madeleine and others},
  journal={Nature methods},
  volume={16},
  number={1},
  pages={111--116},
  year={2019},
  publisher={Nature Publishing Group US New York}
}

@article{tong2022videomae,
  title={Videomae: Masked autoencoders are data-efficient learners for self-supervised video pre-training},
  author={Tong, Zhan and Song, Yibing and Wang, Jue and Wang, Limin},
  journal={Advances in neural information processing systems},
  volume={35},
  pages={10078--10093},
  year={2022}
}

@article{hsu2021hubert,
  title={Hubert: Self-supervised speech representation learning by masked prediction of hidden units},
  author={Hsu, Wei-Ning and Bolte, Benjamin and Tsai, Yao-Hung Hubert and Lakhotia, Kushal and Salakhutdinov, Ruslan and Mohamed, Abdelrahman},
  journal={IEEE/ACM transactions on audio, speech, and language processing},
  volume={29},
  pages={3451--3460},
  year={2021},
  publisher={IEEE}
}

@article{bai2023qwen,
  title={Qwen technical report},
  author={Bai, Jinze and Bai, Shuai and Chu, Yunfei and Cui, Zeyu and Dang, Kai and Deng, Xiaodong and Fan, Yang and Ge, Wenbin and Han, Yu and Huang, Fei and others},
  journal={arXiv preprint arXiv:2309.16609},
  year={2023}
}

@inproceedings{xu2023bridgetower,
  title={Bridgetower: Building bridges between encoders in vision-language representation learning},
  author={Xu, Xiao and Wu, Chenfei and Rosenman, Shachar and Lal, Vasudev and Che, Wanxiang and Duan, Nan},
  booktitle={Proceedings of the AAAI Conference on Artificial Intelligence},
  volume={37},
  number={9},
  pages={10637--10647},
  year={2023}
}

@article{zacks2007event,
  title={Event perception: a mind-brain perspective.},
  author={Zacks, Jeffrey M and Speer, Nicole K and Swallow, Khena M and Braver, Todd S and Reynolds, Jeremy R},
  journal={Psychological bulletin},
  volume={133},
  number={2},
  pages={273},
  year={2007},
  publisher={American Psychological Association}
}

@article{handwerker2004variation,
  title={Variation of BOLD hemodynamic responses across subjects and brain regions and their effects on statistical analyses},
  author={Handwerker, Daniel A and Ollinger, John M and D'Esposito, Mark},
  journal={Neuroimage},
  volume={21},
  number={4},
  pages={1639--1651},
  year={2004},
  publisher={Elsevier}
}

@article{mueller2013individual,
  title={Individual variability in functional connectivity architecture of the human brain},
  author={Mueller, Sophia and Wang, Danhong and Fox, Michael D and Yeo, BT Thomas and Sepulcre, Jorge and Sabuncu, Mert R and Shafee, Rebecca and Lu, Jie and Liu, Hesheng},
  journal={Neuron},
  volume={77},
  number={3},
  pages={586--595},
  year={2013},
  publisher={Elsevier}
}

@article{sonkusare2019naturalistic,
  title={Naturalistic stimuli in neuroscience: critically acclaimed},
  author={Sonkusare, Saurabh and Breakspear, Michael and Guo, Christine},
  journal={Trends in cognitive sciences},
  volume={23},
  number={8},
  pages={699--714},
  year={2019},
  publisher={Elsevier}
}

@article{finn2022naturalistic,
  title={Naturalistic imaging: The use of ecologically valid conditions to study brain function},
  author={Finn, Emily S and Glerean, Enrico and Hasson, Uri and Vanderwal, Tamara},
  journal={NeuroImage},
  volume={247},
  pages={118776},
  year={2022},
  publisher={Elsevier}
}

@article{khosla2021cortical,
  title={Cortical response to naturalistic stimuli is largely predictable with deep neural networks},
  author={Khosla, Meenakshi and Ngo, Gia H and Jamison, Keith and Kuceyeski, Amy and Sabuncu, Mert R},
  journal={Science Advances},
  volume={7},
  number={22},
  pages={eabe7547},
  year={2021},
  publisher={American Association for the Advancement of Science}
}

@article{oota2025multi,
  title={Multi-modal brain encoding models for multi-modal stimuli},
  author={Oota, Subba Reddy and Pahwa, Khushbu and Marreddy, Mounika and Singh, Maneesh and Gupta, Manish and Raju, Bapi S},
  journal={arXiv preprint arXiv:2505.20027},
  year={2025}
}

@article{tang2023brain,
  title={Brain encoding models based on multimodal transformers can transfer across language and vision},
  author={Tang, Jerry and Du, Meng and Vo, Vy and Lal, Vasudev and Huth, Alexander},
  journal={Advances in neural information processing systems},
  volume={36},
  pages={29654--29666},
  year={2023}
}

@inproceedings{dong2023modeling,
  title={Modeling brain responses to video stimuli using multimodal video transformers},
  author={Dong, Tianai and Toneva, Mariya},
  booktitle={the Conference on Cognitive Computational Neuroscience (CCN 2023)},
  pages={194--197},
  year={2023}
}

@article{takagi2025cross,
  title={Cross-modal deep generative models reveal the cortical representation of dancing},
  author={Takagi, Yu and Shimizu, Daichi and Wakabayashi, Mina and Ohata, Ryu and Imamizu, Hiroshi},
  journal={Nature Communications},
  volume={16},
  number={1},
  pages={9937},
  year={2025},
  publisher={Nature Publishing Group UK London}
}

@article{li2024enhancing,
  title={Enhancing neural encoding models for naturalistic perception with a multi-level integration of deep neural networks and cortical networks},
  author={Li, Yuanning and Yang, Huzheng and Gu, Shi},
  journal={Science Bulletin},
  volume={69},
  number={11},
  pages={1738--1747},
  year={2024},
  publisher={Elsevier}
}

@article{hanke2016studyforrest,
  title={A studyforrest extension, simultaneous fMRI and eye gaze recordings during prolonged natural stimulation},
  author={Hanke, Michael and Adelh{\"o}fer, Nico and Kottke, Daniel and Iacovella, Vittorio and Sengupta, Ayan and Kaule, Falko R and Nigbur, Roland and Waite, Alexander Q and Baumgartner, Florian and Stadler, J{\"o}rg},
  journal={Scientific data},
  volume={3},
  number={1},
  pages={160092},
  year={2016},
  publisher={Nature Publishing Group}
}

@article{pinho2024individual,
  title={Individual Brain Charting dataset extension, third release for movie watching and retinotopy data},
  author={Pinho, Ana Lu{\'\i}sa and Richard, Hugo and Ponce, Ana Fernanda and Eickenberg, Michael and Amadon, Alexis and Dohmatob, Elvis and Denghien, Isabelle and Torre, Juan Jes{\'u}s and Shankar, Swetha and Aggarwal, Himanshu and others},
  journal={Scientific Data},
  volume={11},
  number={1},
  pages={590},
  year={2024},
  publisher={Nature Publishing Group UK London}
}

@article{sutskever2014sequence,
  title={Sequence to sequence learning with neural networks},
  author={Sutskever, Ilya and Vinyals, Oriol and Le, Quoc V},
  journal={Advances in neural information processing systems},
  volume={27},
  year={2014}
}

@article{feilong2023individualized,
  title={The individualized neural tuning model: Precise and generalizable cartography of functional architecture in individual brains},
  author={Feilong, Ma and Nastase, Samuel A and Jiahui, Guo and Halchenko, Yaroslav O and Gobbini, M Ida and Haxby, James V},
  journal={Imaging Neuroscience},
  volume={1},
  pages={1--34},
  year={2023},
  publisher={MIT Press One Broadway, 12th Floor, Cambridge, Massachusetts 02142, USA~…}
}

@inproceedings{vaidya2022self,
  title={Self-supervised models of audio effectively explain human cortical responses to speech},
  author={Vaidya, Aditya R and Jain, Shailee and Huth, Alexander G},
  booktitle={Proceedings of the 39th International Conference on Machine Learning},
  series={Proceedings of Machine Learning Research},
  volume={162},
  pages={21927--21944},
  year={2022},
  publisher={PMLR}
}

@article{bonnasse2024fmri,
  title={fMRI predictors based on language models of increasing complexity recover brain left lateralization},
  author={Bonnasse-Gahot, Laurent and Pallier, Christophe},
  journal={Advances in Neural Information Processing Systems},
  volume={37},
  pages={125231--125263},
  year={2024}
}

@article{vu2011encoding,
  title={Encoding and decoding V1 fMRI responses to natural images with sparse nonparametric models},
  author={Vu, Vincent Q and Ravikumar, Pradeep and Naselaris, Thomas and Kay, Kendrick N and Gallant, Jack L and Yu, Bin},
  journal={The annals of applied statistics},
  volume={5},
  number={2B},
  pages={1159},
  year={2011}
}

@article{tang2025semantic,
  title={Semantic language decoding across participants and stimulus modalities},
  author={Tang, Jerry and Huth, Alexander G},
  journal={Current Biology},
  volume={35},
  number={5},
  pages={1023--1032},
  year={2025},
  publisher={Elsevier}
}

@article{ren2017inter,
  title={Inter-subject functional correlation reveal a hierarchical organization of extrinsic and intrinsic systems in the brain},
  author={Ren, Yudan and Nguyen, Vinh Thai and Guo, Lei and Guo, Christine Cong},
  journal={Scientific reports},
  volume={7},
  number={1},
  pages={10876},
  year={2017},
  publisher={Nature Publishing Group UK London}
}

@article{vanderwal2017individual,
  title={Individual differences in functional connectivity during naturalistic viewing conditions},
  author={Vanderwal, Tamara and Eilbott, Jeffrey and Finn, Emily S and Craddock, R Cameron and Turnbull, Adam and Castellanos, F Xavier},
  journal={NeuroImage},
  volume={157},
  pages={521--530},
  year={2017},
  publisher={Elsevier}
}

@article{yeshurun2021default,
  title={The default mode network: where the idiosyncratic self meets the shared social world},
  author={Yeshurun, Yaara and Nguyen, Mai and Hasson, Uri},
  journal={Nature reviews neuroscience},
  volume={22},
  number={3},
  pages={181--192},
  year={2021},
  publisher={Nature Publishing Group UK London}
}

@article{song2025neural,
  title={A neural network with episodic memory learns causal relationships between narrative events},
  author={Song, Hayoung and Lu, Qihong and Nguyen, Tan T and Chen, Janice and Leong, Yuan Chang and Rosenberg, Monica D and Ching, ShiNung and Zacks, Jeffrey M},
  journal={bioRxiv},
  pages={2025.09.01.673596},
  year={2025},
  publisher={Cold Spring Harbor Laboratory}
}

@article{jain2018incorporating,
  title={Incorporating context into language encoding models for fMRI},
  author={Jain, Shailee and Huth, Alexander},
  journal={Advances in neural information processing systems},
  volume={31},
  year={2018}
}

@article{paugam2024benchmark,
  title={A benchmark of individual auto-regressive models in a massive fMRI dataset},
  author={Paugam, Fran{\c{c}}ois and Pinsard, Basile and Lajoie, Guillaume and Bellec, Pierre},
  journal={Imaging Neuroscience},
  volume={2},
  pages={1--23},
  year={2024},
  publisher={MIT Press 255 Main Street, 9th Floor, Cambridge, Massachusetts 02142, USA~…}
}

@article{jain2020interpretable,
  title={Interpretable multi-timescale models for predicting fMRI responses to continuous natural speech},
  author={Jain, Shailee and Vo, Vy and Mahto, Shivangi and LeBel, Amanda and Turek, Javier S and Huth, Alexander},
  journal={Advances in Neural Information Processing Systems},
  volume={33},
  pages={13738--13749},
  year={2020}
}

@article{jindal2026does,
  title={How does longer temporal context enhance multimodal narrative video processing in the brain?},
  author={Jindal, Prachi and Khandelwal, Anant and Gupta, Manish and Raju, Bapi S and Oota, Subba Reddy and Chakraborty, Tanmoy},
  journal={arXiv preprint arXiv:2602.07570},
  year={2026}
}

@article{tang2026rapid,
  title={Rapid cortical mapping with cross-participant encoding models},
  author={Tang, Jerry and Huth, Alexander G},
  journal={Imaging Neuroscience},
  volume={4},
  pages={IMAG.a.1349},
  year={2026},
  publisher={MIT Press 255 Main Street, 9th Floor, Cambridge, Massachusetts 02142, USA~…}
}

\markboth{SUPPLEMENTARY MATERIAL}{SUPPLEMENTARY MATERIAL}
\section*{Supplementary Material}

\setcounter{figure}{0}
\renewcommand{\thefigure}{S\arabic{figure}}
\setcounter{table}{0}
\renewcommand{\thetable}{S\arabic{table}}

\providecommand{\theHfigure}{\thefigure}
\providecommand{\theHtable}{\thetable}
\renewcommand{\theHfigure}{S\arabic{figure}}
\renewcommand{\theHtable}{S\arabic{table}}

\begin{figure}[H]
\begin{center}
\includegraphics[page=3,width=0.85\textwidth]{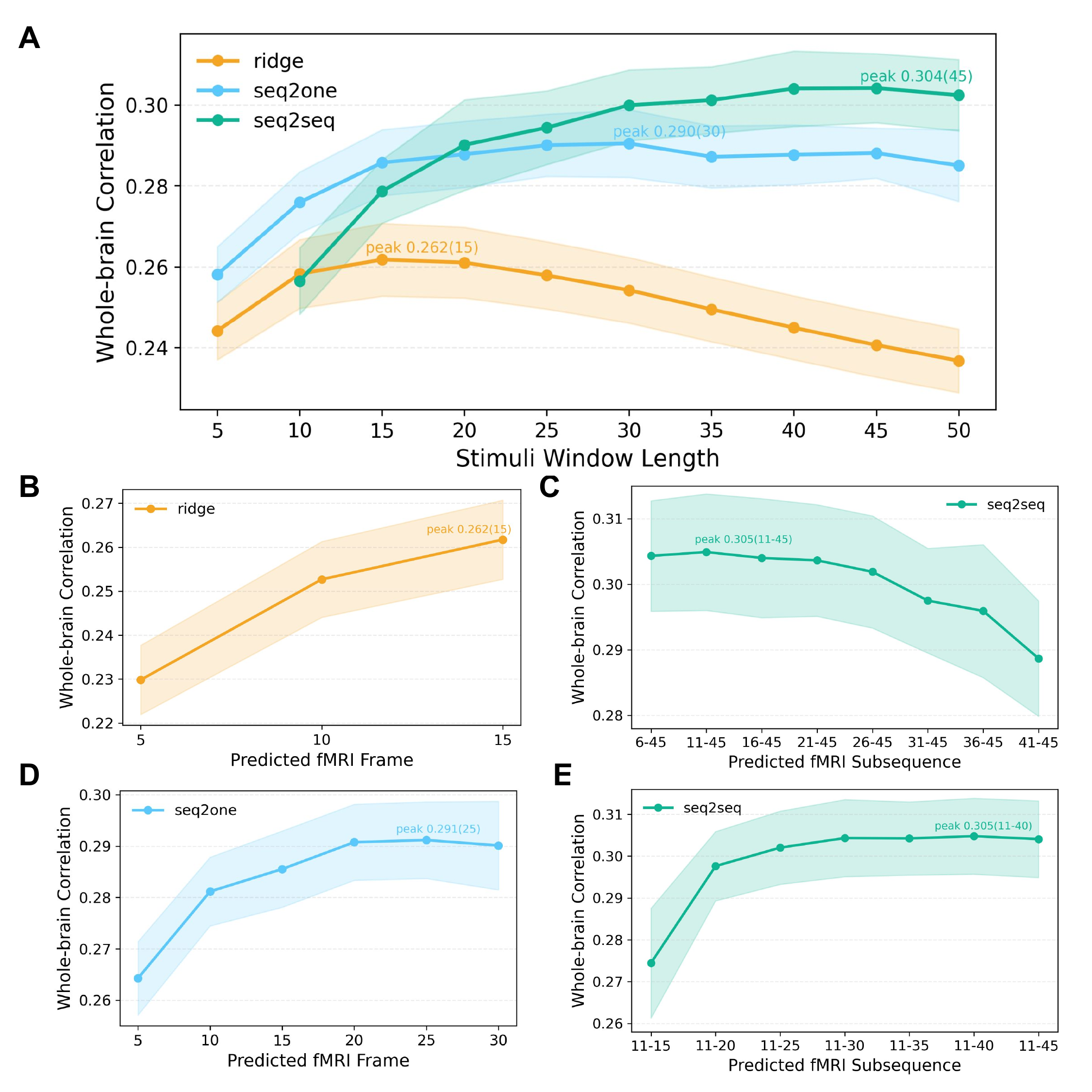}
\caption{\textbf{Temporal model selection on the validation split.}
\textbf{(A)} Whole-brain prediction accuracy as a function of the stimulus context length \(k\) for ridge (orange), transformer seq2one (blue), and transformer seq2seq (green). Solid lines denote subject means (n=4) and shaded bands indicate \(\pm\) s.e.m. The sweep identifies \(k=15\) (ridge), \(k=30\) (seq2one), and \(k=45\) (seq2seq).
\textbf{(B–C)} With the stimulus window length fixed at \(k\), ridge (B) and seq2one (C) perform a target sweep over the predicted fMRI frame in 5~TR increments from TR~5 to TR~\(k\) with a 1~TR sliding-window stride. The optima occur at frame~15 and frame~25, respectively.
\textbf{(D–E)} For seq2seq with \(k=45\), we evaluated output subsequences with a minimum prediction start of  \(\ge 5\)~TRs: (D) the start \(a\) varies with the end \(b\) fixed. (E) the end \(b\) varies with the start \(a\) fixed. Both sweeps exhibit a broad performance plateau when the output segment includes late TRs near the window end. Balancing accuracy and robustness, the seq2seq output is fixed to the contiguous sequence TRs~11-40.}

\label{sup_1}
\end{center}
\end{figure}

\begin{figure}[H]
\begin{center}
\includegraphics[
    page=1,
    width=1.04\textwidth,
    trim=0cm 3cm 0cm 0cm,
    clip
]{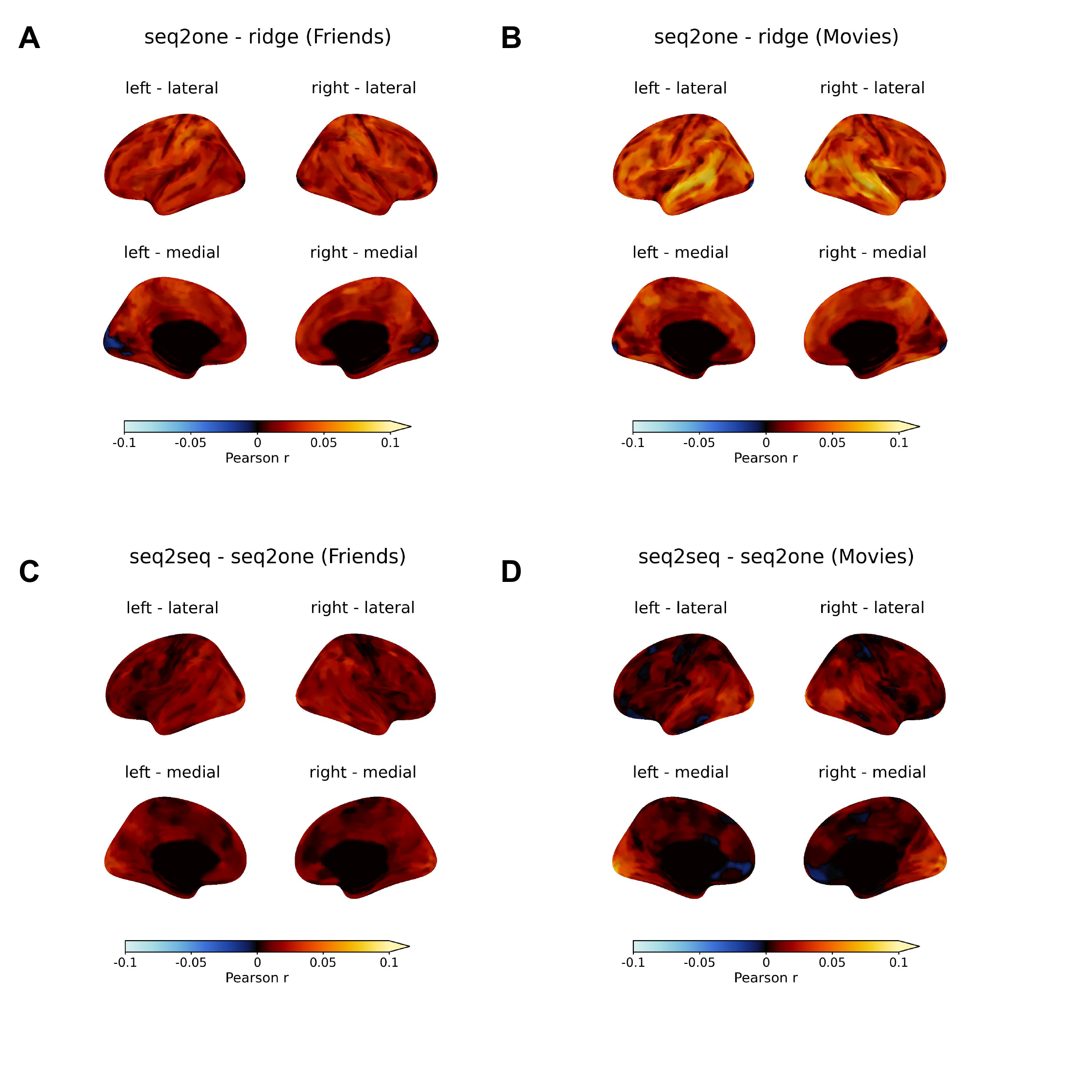}
\caption{\textbf{(A–B)} Difference maps (seq2one$-$ridge) showing widespread improvements across cortex. \textbf{(C–D)} Difference maps (seq2seq$-$seq2one) showing further gains in association regions, with only minor negative effects (blue) in limited parcels.}
\label{sup_2}
\end{center}
\end{figure}

\begin{figure}[H]
\begin{center}
\includegraphics[
    page=2,
    width=1.04\textwidth,
    trim=0cm 21cm 0cm 0cm,
    clip
]{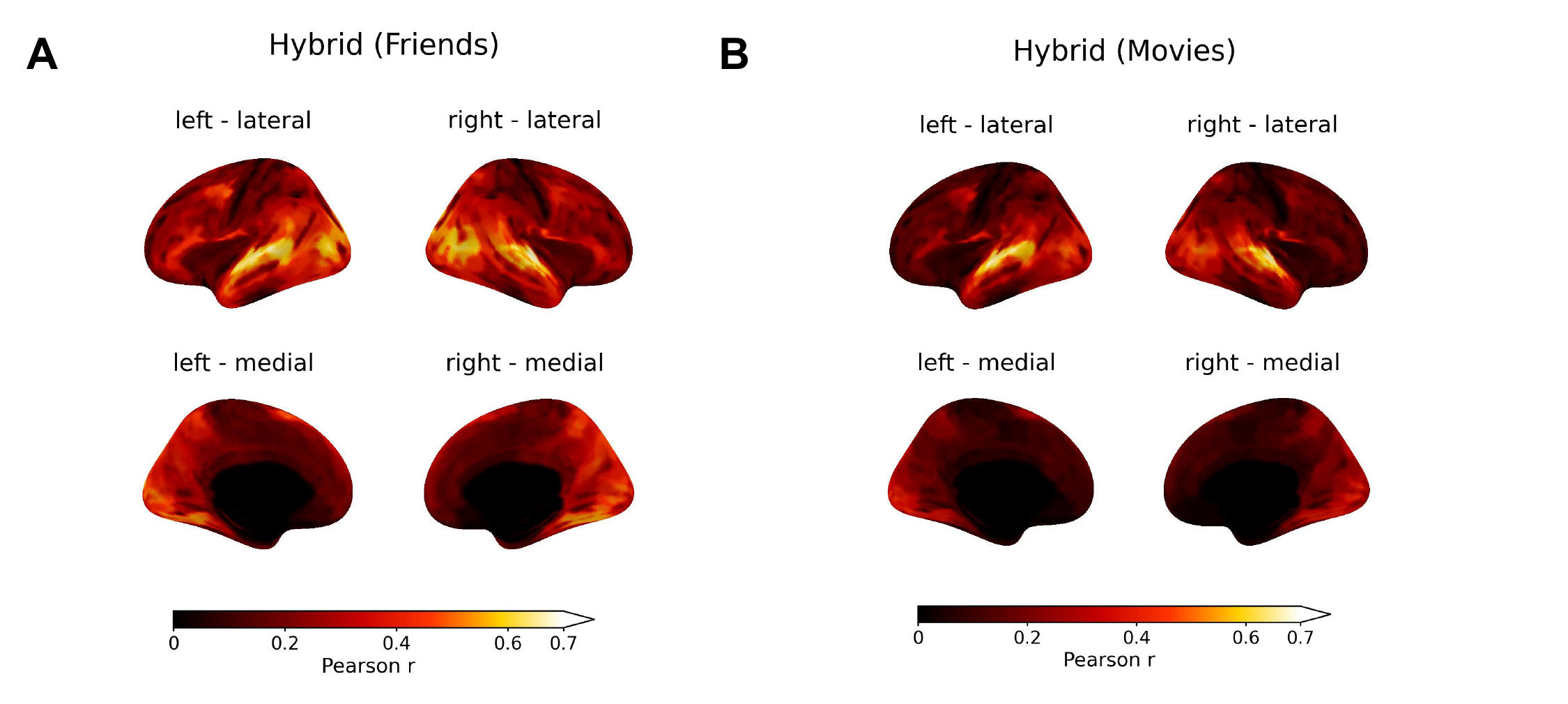}
\caption{\textbf{(A–B)} Whole-brain prediction maps for hybrid on Friends and Movies.}
\label{sup_3}
\end{center}
\end{figure}

\begin{table}[H]
\begin{threeparttable}
\caption{\textbf{Per-subject test correlations for temporal models on held-out data.}}
\label{tab:s1}
\setlength{\tabcolsep}{10pt}%
\renewcommand{\arraystretch}{1.15}%
\centering

\begin{tabular*}{\textwidth}{@{\extracolsep{\fill}}lccccc@{}}
\toprule
 & \multicolumn{4}{c}{\textbf{Friends (test)}} & \textbf{Mean $\pm$ s.e.m.} \\
\cmidrule(lr){2-5}
\textbf{Model} & sub-01 & sub-02 & sub-03 & sub-05 &  \\
\midrule
Ridge   & 0.229 & 0.261 & 0.286 & 0.223 & 0.250 $\pm$ 0.015 \\
Seq2one & 0.252 & 0.296 & 0.312 & 0.250 & 0.278 $\pm$ 0.016 \\
Seq2seq & 0.268 & 0.309 & 0.331 & 0.264 & 0.293 $\pm$ 0.016 \\
\midrule
 & \multicolumn{4}{c}{\textbf{Movies (test)}} & \textbf{Mean $\pm$ s.e.m.} \\
\cmidrule(lr){2-5}
\textbf{Model} & sub-01 & sub-02 & sub-03 & sub-05 &  \\
\midrule
Ridge   & 0.163 & 0.146 & 0.171 & 0.123 & 0.151 $\pm$ 0.010 \\
Seq2one & 0.202 & 0.184 & 0.207 & 0.160 & 0.188 $\pm$ 0.011 \\
Seq2seq & 0.214 & 0.198 & 0.220 & 0.171 & 0.201 $\pm$ 0.011 \\
\bottomrule
\end{tabular*}

\begin{tablenotes}[flushleft]\footnotesize
\item Correlations were Fisher-transformed before averaging across parcels and, for movie analyses, across films. Group means were computed in Fisher $z$ space and inverse-transformed to Pearson $r$, while s.e.m. values were computed across participant-level Pearson $r$ values.
\end{tablenotes}
\end{threeparttable}
\end{table}

\begin{table}[H]
\begin{threeparttable}
\caption{\textbf{Per-subject test correlations for shared, individual, and hybrid parameterization strategies on held-out data.}}
\label{tab:s2}
\setlength{\tabcolsep}{10pt}%
\renewcommand{\arraystretch}{1.15}%
\centering

\begin{tabular*}{\textwidth}{@{\extracolsep{\fill}}lccccc@{}}
\toprule
 & \multicolumn{4}{c}{\textbf{Friends (test)}} & \textbf{Mean $\pm$ s.e.m.} \\
\cmidrule(lr){2-5}
\textbf{Model} & sub-01 & sub-02 & sub-03 & sub-05 &  \\
\midrule
Shared      & 0.257 & 0.286 & 0.289 & 0.253 & 0.271 $\pm$ 0.009 \\
Individual  & 0.268 & 0.309 & 0.330 & 0.264 & 0.293 $\pm$ 0.016 \\
Hybrid      & 0.303 & 0.313 & 0.337 & 0.287 & 0.310 $\pm$ 0.011 \\
\midrule
 & \multicolumn{4}{c}{\textbf{Movies (test)}} & \textbf{Mean $\pm$ s.e.m.} \\
\cmidrule(lr){2-5}
\textbf{Model} & sub-01 & sub-02 & sub-03 & sub-05 &  \\
\midrule
Shared      & 0.192 & 0.190 & 0.196 & 0.166 & 0.186 $\pm$ 0.007 \\
Individual  & 0.214 & 0.198 & 0.221 & 0.173 & 0.202 $\pm$ 0.011 \\
Hybrid      & 0.222 & 0.204 & 0.231 & 0.184 & 0.210 $\pm$ 0.011 \\
\bottomrule
\end{tabular*}

\begin{tablenotes}[flushleft]\footnotesize
\item Correlations were Fisher-transformed before averaging across parcels and, for movie analyses, across films. Group means were computed in Fisher $z$ space and inverse-transformed to Pearson $r$, while s.e.m. values were computed across participant-level Pearson $r$ values.
\end{tablenotes}
\end{threeparttable}
\end{table}

\begin{table}[H]
\centering
\small
\setlength{\tabcolsep}{6pt}
\caption{\textbf{Film-wise transfer performance in the small-in scenario.}}
\begin{tabular}{llccccc}
\toprule
Film & Exp & Overall & Sub1 & Sub2 & Sub3 & Sub5 \\
\midrule
Bourne
& Individual & 0.1363 & $0.1574\pm0.0282$ & $0.1038\pm0.0293$ & $0.1680\pm0.0186$ & $0.1156\pm0.0198$ \\
& Hybrid     & 0.1696 & $0.1909\pm0.0339$ & $0.1422\pm0.0395$ & $0.1934\pm0.0210$ & $0.1515\pm0.0315$ \\
& Shared     & 0.1638 & $0.1839\pm0.0347$ & $0.1361\pm0.0339$ & $0.1875\pm0.0157$ & $0.1474\pm0.0326$ \\
\midrule
Figures
& Individual & 0.1721 & $0.1878\pm0.0187$ & $0.1800\pm0.0193$ & $0.1770\pm0.0161$ & $0.1435\pm0.0270$ \\
& Hybrid     & 0.2013 & $0.2260\pm0.0227$ & $0.2064\pm0.0138$ & $0.1981\pm0.0110$ & $0.1745\pm0.0295$ \\
& Shared     & 0.1888 & $0.2096\pm0.0173$ & $0.1934\pm0.0145$ & $0.1867\pm0.0153$ & $0.1655\pm0.0258$ \\
\midrule
Life
& Individual & 0.1219 & $0.1316\pm0.0396$ & $0.1349\pm0.0390$ & $0.1687\pm0.0236$ & $0.0517\pm0.0544$ \\
& Hybrid     & 0.1333 & $0.1290\pm0.0747$ & $0.1342\pm0.0759$ & $0.1527\pm0.0761$ & $0.1172\pm0.0452$ \\
& Shared     & 0.1163 & $0.1221\pm0.0666$ & $0.1178\pm0.0711$ & $0.1251\pm0.0611$ & $0.1003\pm0.0585$ \\
\midrule
Wolf
& Individual & 0.1698 & $0.1866\pm0.0105$ & $0.1703\pm0.0089$ & $0.1664\pm0.0214$ & $0.1560\pm0.0044$ \\
& Hybrid     & 0.2008 & $0.2136\pm0.0103$ & $0.1986\pm0.0106$ & $0.2024\pm0.0276$ & $0.1886\pm0.0059$ \\
& Shared     & 0.1871 & $0.1977\pm0.0072$ & $0.1870\pm0.0104$ & $0.1869\pm0.0264$ & $0.1767\pm0.0044$ \\
\bottomrule
\end{tabular}

\label{tab:s3}
\begin{tablenotes}[flushleft]\footnotesize
\item Correlations were Fisher-transformed before any averaging across parcels and subjects. Means and s.e.m. were computed in Fisher $z$ space across cross-validation folds and then inverse-transformed to Pearson $r$. All reported values are Pearson $r$.

\end{tablenotes}
\end{table}

\end{document}